\documentclass[12pt]{article}

\usepackage[square,numbers]{natbib}
\usepackage{graphicx}
\usepackage{amsfonts}
\usepackage{algorithmic}
\usepackage{verbatim}
\usepackage{mathtools}
\usepackage{graphicx}
\usepackage{tikz}
\usepackage{stmaryrd}
\usepackage{booktabs}
\usepackage{amssymb}
\usepackage{amsmath}
\usetikzlibrary{positioning}
\usepackage{caption}
\usepackage{subcaption}
\usepackage{diagbox}

\usepackage{soul}

\usetikzlibrary{matrix,chains,positioning,decorations.pathreplacing,arrows}

\newtheorem{algo}{Algorithm}{\bf}{\it}
\DeclareMathOperator*{\argmax}{arg\,max}

\title{
	Strategic bidding in freight transport using deep reinforcement learning}
\author{W.J.A. van Heeswijk}

\begin{document}
	\maketitle

\begin{abstract}
This paper presents a multi-agent reinforcement learning algorithm to represent strategic bidding behavior in freight transport markets. Using this algorithm, we investigate whether feasible market equilibriums arise without any  central control or communication between agents. Studying behavior in such environments may serve as a stepping stone towards self-organizing logistics systems like the Physical Internet. We model an agent-based environment in which a shipper and a carrier actively learn bidding strategies using policy gradient methods, posing bid- and ask prices at the individual container level. Both agents aim to learn the best response given the expected behavior of the opposing agent. A neutral broker  allocates jobs based on  bid-ask spreads. 

Our game-theoretical analysis and numerical experiments focus on behavioral insights. To evaluate system performance, we measure adherence to Nash equilibria, fairness of reward division and utilization of transport capacity. We observe good performance both in predictable, deterministic settings ($\sim$95\% adherence to Nash equilibria) and highly stochastic environments ($\sim$85\% adherence). Risk-seeking behavior may increase an agent's reward share, as long as the strategies are not overly aggressive. The results suggest a  potential for full automation and decentralization of freight transport markets.
\end{abstract}

\section{Introduction}\label{sec:introduction}
While the transport sector traditionally relies on fixed transport contracts and manual negotiations, modern technology allows for automated negotiation and more appropriate responses to the inherently dynamic nature of freight transport. Paradigms such as the Physical Internet and self-organizing logistics provide conceptual outlines for the organization of such systems, yet many of their facets remain unexplored. This paper investigates a strategic bidding mechanism based on multi-agent reinforcement learning, deliberately exploring a setting without any communication or centralized control, presenting decentralized planning in its pure form. 

We study a multi-agent environment containing a carrier, shipper and broker. For each individual transport job (e.g., a smart container), the shipper and carrier pose bid- and ask prices respectively. The neutral broker agent -- with a business model inspired by financial markets -- matches these bids and asks by solving a knapsack problem that maximizes bid-ask spread. The shipper and carrier actively learn strategies that aim to maximize their own reward given the expected strategy of the opponent. As such, both agents may constantly adjust their strategy in an online learning setting. The aim of this setup is to represent a self-organizing freight transport market. The behavior of such a market is the key interest of this work, Inspired by game theory, efficiency and fairness are the primary performance metrics.

This paper expands upon Van Heeswijk \cite{vanheeswijk2020b} -- which introduced a learning bidding agent -- in several ways. First, the carrier, which was a passive price-taker in the earlier work, now is a learning agent with a dynamic strategy. This vastly increases the complexity and implications of the system, introducing the strategic bidding dimension. Second, we add a broker agent to the setting, exploring the possible role of a neutral transport planner in self-organizing systems. Third, we test a variety of actor-critic models and deep learning techniques, whereas the earlier work relies on a basic policy gradient algorithm. Fourth, we root the performance metrics and numerical results in game-theoretical foundations.

The remainder of this paper is structured as follows. Section~\ref{sec:literature} discusses related literature. In Section~\ref{sec:problem_description} we define the system as a Markov Decision Problem (MDP) model. Section~\ref{sec:solution_method} presents the solution method; we describe several variants of policy gradient algorithms to learn bidding- and asking prices. Section~\ref{sec:experimental_design} presents the experimental design and Section~\ref{sec:numerical_results} the numerical results. Section~\ref{sec:conclusions} ends the paper with the main conclusions.

\section{Literature review}\label{sec:literature}
 This literature overview is structured as follows. First, we discuss the core concepts of self-organizing logistics and the Physical Internet. Second, we highlight several studies on reinforcement learning in the Delivery Dispatching Problem, which relates to our setting. Third, we discuss studies that address the topic of bidding in freight transport. Fourth, we assess the links with game theory, which will be used for our analysis and experimental design.
 
 The inspiration for this paper stems from the Physical Internet paradigm.  We refer to the seminal works of Montreuil \cite{montreuil2011,montreuil2013} for a conceptual outline of the Physical Internet, thoroughly addressing its foundations. The Physical Internet envisions an open market at which logistics services are offered, with automated interactions between smart containers and other constituents of the Physical Internet determining routes and schedules. Sallez \textit{et al.} \cite{sallez2016} emphasize the active role that smart containers have, being able to communicate, memorize, negotiate, and learn both individually and collectively. Ambra \textit{et al.} \cite{ambra2019} present a recent literature review of work performed in the domain of the Physical Internet. Interestingly, their overview does not mention any paper that defines the smart container itself as a key actor. Instead, existing works seem to focus on traditional actors such as carriers, shippers and logistics service providers, even though smart containers supposedly route themselves in the Physical Internet.
 
 The problem studied in this paper is related to the Delivery Dispatching Problem \cite{minkoff1993}, which entails dispatching decisions from a carrier's perspective. In this problem setting, transport jobs arrive at a hub according to some external stochastic process. The carrier subsequently decides which subset of jobs to accept, also anticipating future jobs that arrive according to the stochastic process. The most basic instances may be solved with queuing models, but more complicated variants quickly become computationally intractable, such that researchers often resort to reinforcement learning to learn high-quality strategies. We highlight some recent works in this domain. Klapp \textit{et al.} \cite{klapp2018} develop an algorithm that solves the dispatching problem for a transport service operating on the real line. Van Heeswijk \& La Poutr{\'e} \cite{vanheeswijk2018b} compare centralized and decentralized transport for networks with fixed line transport services, concluding that decentralized planning yields considerable computational benefits. Van Heeswijk \textit{et al.} \cite{vanheeswijk2015,vanheeswijk2019,vanheeswijk2020c} study several variants of the DDP -- including a routing component and a multi-agent setting -- and use value function approximation to find strategies. Voccia \textit{et al.} \cite{voccia2019} solve a variant that includes both pickups and deliveries. Our current paper distinguishes itself from the aforementioned works by assigning jobs based on bid-ask spread (neutral perspective) rather than transport efficiency (carrier perspective).
 
 Next, we highlight related works on optimal bidding in freight transport; most of these studies seem to adopt a viewpoint in which competing carriers bid on transport jobs. For instance, Yan \textit{et al.} \cite{yan2018} propose a particle swarm optimization algorithm used by carriers to place bids on jobs. Miller \& Nie \cite{miller2019} present a solution that emphasizes the importance of integrating carrier competition, routing and bidding. Wang \textit{et al.} \cite{wang2018} design a reinforcement learning algorithm based on knowledge gradients to solve for a bidding structure with a broker intermediating between carriers and shippers. The broker aims to propose a price that satisfies both carrier and shipper, taking a percentage of accepted bids as its reward.  In a Physical Internet context, Qiao \textit{et al.} \cite{qiao2019} model hubs as spot freight markets where carriers can place bids on transport bids. To this end, they propose a dynamic pricing model based on an auction mechanism. Most studies assume that shippers have limited to no influence in the bidding process; we aim to add a fresh perspective with this work. This paper builds onto the work of Van Heeswijk \cite{vanheeswijk2020b}, in which only the shipper is a learning agent and the carrier is a passive price taker. The author uses a policy gradient algorithm to learn the bidding strategy. To the best of our knowledge, there are no comprehensive studies that explicitly model both carriers and shippers as intelligent bidding agents.

The experimental results in this study will be analyzed from a game theoretical perspective. Conceptually, the bid-ask problem may be classified as an infinitely repeated non-cooperative game, in which both agents aim to maximize their average reward. More specifically, it may be classified as a bargaining game such as defined by Nash \cite{nash1953}, in which both agents claim a share of a system-wide gain. Folk theorems provide insights on equilibria in such settings \cite{friedman1971}. They state that each payoff profile that is both feasible and individually rational in the one-shot game constitutes a Nash equilibrium in the repeated game. For bargaining games, the presence of a threat or disagreement point for deviating opponents is essential to prove the existence of Nash equilibria. Aumann \& Shapley \cite{aumann1994} and Rubinstein \cite{rubinstein1980,rubinstein1994} present solutions for Nash equilibria under temporary punishments. The Nash equilibrium is a clear performance metric for the efficiency of our transport market. 

\section{System description}\label{sec:problem_description}

This section formally defines our system in the form of a Markov Decision Process model. In Section~\ref{ssec:model_outline} we provide a high-level outline of the system and the agents involved. Section~\ref{ssec:state} describes the system state; Section~\ref{ssec:decisions} follows up with the decisions and reward functions. Finally, Section~\ref{ssec:transition_function} provides the transition function to complete the model definition.

\subsection{Model outline}\label{ssec:model_outline}
The system contains three agent types: (i) the shipper ($S$), (ii) the carrier ($C$) and (iii) the broker ($B$). We provide a global outline here; more detail follows in the subsequent sections.

\begin{figure}
	\includegraphics[width=\textwidth]{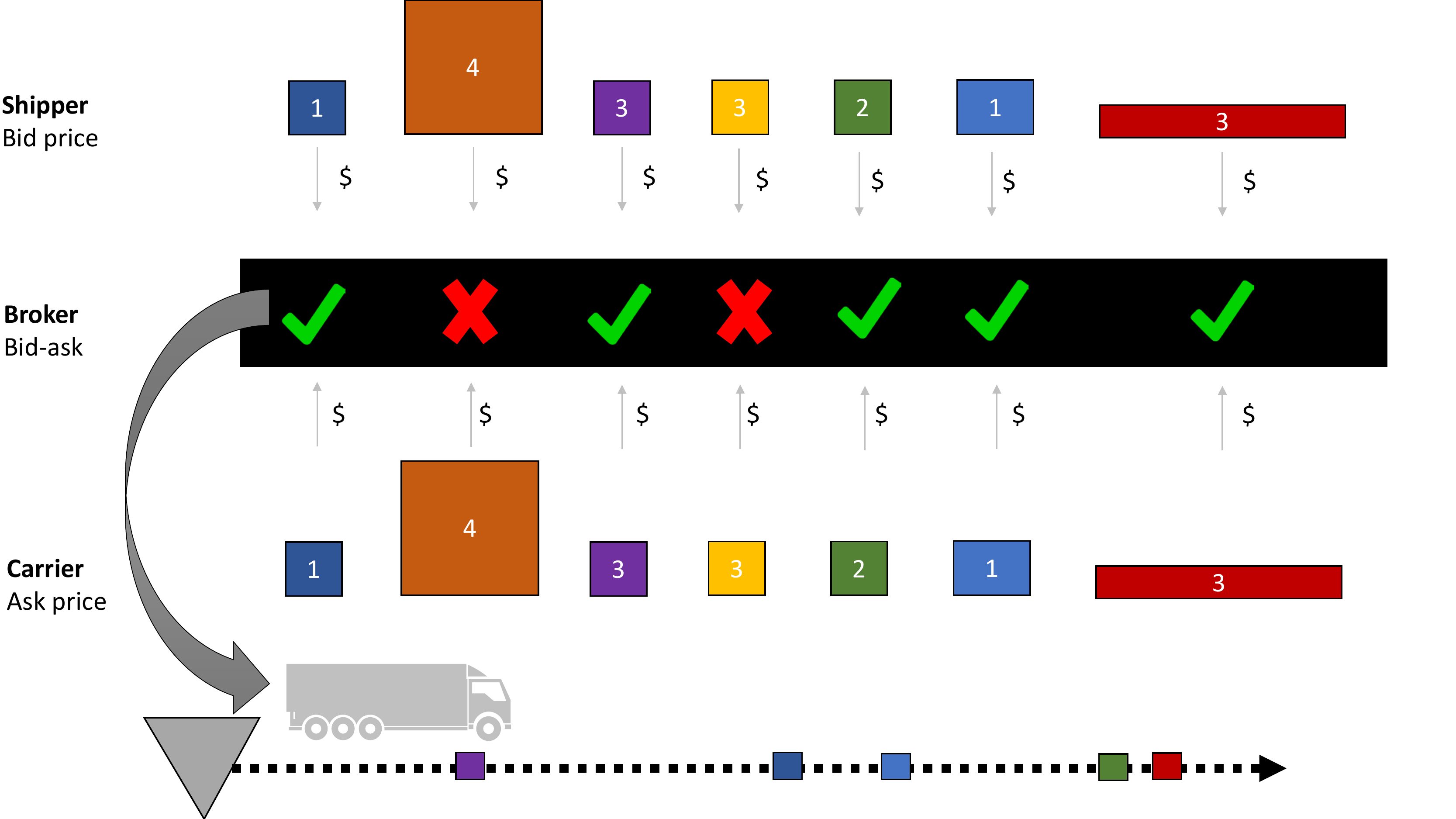}
	\caption{Visual representation of the bid-ask system. For each job, the shipper and carrier pose a bid- and ask price respectively. The broker assigns jobs based on bid-ask spread. For the carrier and shipper, the process is essentially a black box.} \label{fig:problem_illustration}
\end{figure}

We consider a singular transport service with a fixed origin (e.g., a transport hub). Every day, the \textit{shipper} places individual bids (highlighting the autonomy of smart containers) for each job to be transported to its destination on a real line. The job is shipped if its bid is accepted. If the job reaches its due date and is still not shipped, it is removed from the system as a failed job. Second, the \textit{carrier} is responsible for performing the transport service. Without knowing the bid price, it poses a daily ask price for each job. Depending on volume and distance, each job has its own marginal transport costs. When the job is assigned by the broker, the carrier is obliged to transport it and receives the requested ask price. Third, the freight \textit{broker} is responsible for scheduling jobs. After receiving all bids and asks for the day, the broker assigns transport jobs to the carrier. Its profit is the difference between the bid- and ask price of each job that is shipped. This means that (i) jobs are never shipped when the ask price exceeds the bid price and (ii) in case transport capacity is insufficient, the broker assigns jobs in a way that maximizes its own total profit. We illustrate the strategic bidding problem in Figure~\ref{fig:problem_illustration}. 

We consider an infinite horizon setting in which bidding- and asking strategies are updated each day (i.e., an online learning setting\cite{powell2007}). For practical and notational purposes, we set the length of the time horizon to $N=\infty$ and the corresponding decision epochs (representing a day) to $n \in \{0,1,...,N\}$. Every day $n$, a transport service with fixed capacity $\zeta^{C}$ departs along the real line. Neither past bid- and ask prices nor job allocations impact current decisions, satisfying the Markovian memoryless property.

\subsection{State description}\label{ssec:state}
The system state is defined as the set of transport jobs. An individual job is defined by an attribute vector $\boldsymbol{j}$. In addition to the global time horizon that runs till $N$, each job has an individual time horizon $\mathcal{T}_{\boldsymbol{j}}$ that corresponds to its time till due date (i.e., decreasing with time), with $t \in \mathcal{T}_{\boldsymbol{j}}$ being decision epochs. Each job is represented by an attribute vector: 
\begin{equation}
\boldsymbol{j} = 
\begin{pmatrix}
j_\tau & = & \text{time till due date} \\
j_d & = & \text{distance to destination} \\
j_v & = & \text{container volume} \notag
\end{pmatrix}
\end{equation}

The integer attribute $j_\tau \in [0,\tau^{max}]$ indicates how many decision epochs remain until the latest possible shipment date.  Whenever a new job arrives in the system, we initialize $\mathcal{T}_{\boldsymbol{j}}=\{j_\tau,j_\tau-1,\ldots,0\}$. The time horizon is tied to the individual job; the corresponding time till due date -- represented by attribute $j_\tau$ -- is decremented with each time step. If $j_\tau=0$ and the job has not been shipped, it is removed from the system. Next, the attribute $j_d \in[1,d^{max}]$ indicates the distance between origin and destination. Finally, the job volume $j_v \in [1,\zeta^{max}]$, with $\zeta^{max}\leq\zeta^{C}$  represents the transport capacity required for the job.

The system state at day $n$ is represented by the set $\boldsymbol{J}_{n}$, which contains all jobs $\boldsymbol{j}$ present in the system. We use $\mathcal{J}_{n}$ to denote the set of feasible states.

\subsection{Decisions and rewards}\label{ssec:decisions}
We introduce the decisions and reward functions.  The shipper places a bid $b_{n,\boldsymbol{j}} \in \mathbb{R}$ for each job and the carrier poses an ask price $a_{n,\boldsymbol{j}} \in \mathbb{R}$ for each job. The broker decides which jobs to assign to the carrier, earning $b_{n,\boldsymbol{j}}-a_{n,\boldsymbol{j}}$ for each job that is shipped. 

We first discuss the decision and reward function of the shipper in more detail. The shipper places bids according to a strategy $\pi_n^{shp}=\mathbb{P}(b_{n,\boldsymbol{j}} \mid (\boldsymbol{j},\boldsymbol{J}_n))$, with $\Pi^S$ denoting the set of feasible strategies; how to obtain such a strategy will be explained in Section~\ref{sec:solution_method}. The bids of all jobs are stored in a vector $\boldsymbol{b} = [b_{\boldsymbol{j}}]_{\forall \boldsymbol{j} \in \boldsymbol{J}_{n}}$. Remind that the payoff depends on the decision of the broker and is unknown when posing the bid. The maximum willingness to pay for transporting job $j$ is represented by $ c_{\boldsymbol{j}}^{S,max}=c^{S,max}\cdot j_v \cdot j_d$ (i.e., depending on volume and distance); this value is used to compute the reward. Note failing a job trumps bidding over the maximum willingness. At any given decision epoch, the direct reward function for individual jobs is defined as follows:
 
 \begin{align}\label{eq:rewardshipper}
 r_{\boldsymbol{j}}^{S}(\gamma_{\boldsymbol{j}},b_{\boldsymbol{j}})= 
 \begin{cases}
 c_{\boldsymbol{j}}^{S,max}- b_{\boldsymbol{j}} &  \text{if} \qquad \gamma_{\boldsymbol{j}}=1  \\
0 & \text{otherwise}
 \end{cases}\enspace.
 \end{align}

The carrier makes its decision according to a strategy $\pi_n^{C}=\mathbb{P}(a_{n,\boldsymbol{j}} \mid (\boldsymbol{j},\boldsymbol{J}_n))$, with $\Pi^C$ denoting the set of strategies. Similar to the bids, the ask prices during an episode are stored in a vector $\boldsymbol{a}$. Each job has a marginal transport cost that depends on its distance and volume: $c_{\boldsymbol{j}}^{C,trn}=  c^{C,trn} \cdot j_v \cdot j_d$. Posing an ask price below the transport costs yields a loss when accepted. If a job is assigned, the carrier's reward is the difference between the ask price and the transport costs:

\begin{align}\label{eq:rewardcarrier}
r_{\boldsymbol{j}}^{C}(\gamma_{\boldsymbol{j}},a_{\boldsymbol{j}})= 
\begin{cases}
a_{\boldsymbol{j}}- c_{\boldsymbol{j}}^{C,trn} &  \text{if} \qquad \gamma_{\boldsymbol{j}}=1 \\
0 & \text{otherwise} 
\end{cases}\enspace.
\end{align}

For the broker, the value of a job is its bid price ($b_{\boldsymbol{j}}$) minus its ask price ($a_{\boldsymbol{j}}$). Using these values and the capacity of the carrier, the broker solves a 0-1 knapsack problem -- using dynamic programming \cite{kellerer2004} -- to assign the jobs in a way that maximizes its own profit.  The broker maximizes its direct rewards by selecting $\boldsymbol{\gamma}$ as follows:
\begin{align}\label{eq:selectionbroker}
\argmax_{\boldsymbol{\gamma}\in \Gamma(\boldsymbol{J}_{n})} \left(\sum_{\boldsymbol{j} \in \boldsymbol{J}_{n}} {\gamma}_{\boldsymbol{j}}(b_{\boldsymbol{j}} - a_{\boldsymbol{j}})\right)\enspace,
\end{align}
s.t.
\begin{align}
\sum_{\boldsymbol{j} \in \boldsymbol{J}_{n}} {\gamma}_{\boldsymbol{j}} \cdot j_v \leq \zeta^{C}\enspace. \notag
\end{align}

The corresponding reward function for the broker is:

\begin{align}
r_{\boldsymbol{j}}^{B}(\gamma_{\boldsymbol{j}},b_{\boldsymbol{j}},a_{\boldsymbol{j}})= 
\begin{cases}
b_{\boldsymbol{j}}-a_{\boldsymbol{j}} &  \text{if} \qquad \gamma_{\boldsymbol{j}}=1 \\
0 &\text{otherwise} 
\end{cases}\enspace.
\end{align}

Note that the broker will never select a job which has a higher ask price than a bid price; this would yield a negative payoff. We formalize this minor result in Lemma 1, serving as a building block for later proofs.

\textbf{Lemma 1: Job selection when bid price is lower than ask price}\\
\textit{If $b_{\boldsymbol{j}}<a_{\boldsymbol{j}}$, the broker will always set ${\gamma}_{\boldsymbol{j}}=0$ to maximize its profits.}

\textbf{Proof:}
The proof is found in Appendix A. The broker always rejects jobs that would yield a negative payoff, since rejecting yields a payoff of 0.  $\blacksquare$

\subsection{Transition function}\label{ssec:transition_function}

To conclude the system definition, we describe the transition function for the system state that occurs in the time step from decision epoch $n$ to $n+1$. Two state changes occur during a time step; (i) due dates are decreased for all jobs that are not shipped or failed and (ii) newly arrived jobs are added to the state. We define the set of new jobs arriving at $n+1$  by $\boldsymbol{\tilde{J}}_{n+1} \in \mathcal{\tilde{J}}_{n}$. The transition function $X \colon (\boldsymbol{J}_{n},\boldsymbol{\tilde{J}}_{n+1},\boldsymbol{\gamma}) \mapsto \boldsymbol{J}_{n+1}$ is a sequential procedure that is outlined in Algorithm 1.

\setcounter{algo}{0}
\begin{algo}Transition function $X(\boldsymbol{J}_{n},\boldsymbol{\tilde{J}}_{n+1},\boldsymbol{\gamma}_{n})$
\end{algo}\small
\begin{tabular}{ l  l l}
	\toprule		
	0: & Input: $\boldsymbol{J}_{n},\boldsymbol{\tilde{J}}_{n+1},\boldsymbol{\gamma}_{n}$& $\blacktriangleright$ Current state, job arrivals, shipping selection\\
	1: & $\boldsymbol{J}_{n+1} \mapsfrom \emptyset$ & $\blacktriangleright$  Initialize next state\\	
	2: & $\boldsymbol{J}_{n}^x \mapsfrom \boldsymbol{J}_{n}$& $\blacktriangleright$ Copy state (post-decision state)\\
	3: & $\forall \boldsymbol{j} \in \boldsymbol{J}_{n}$ & $\blacktriangleright$ Loop over all jobs\\
	4: & \qquad $\boldsymbol{J}_{n}^x \mapsfrom  \boldsymbol{J}_{n}^x \setminus \boldsymbol{j}  \mid \gamma_{{n},\boldsymbol{j}}=1$ & $\blacktriangleright$ Remove shipped job\\
	5: & \qquad $\boldsymbol{J}_{n}^x \mapsfrom  \boldsymbol{J}_{n}^x \setminus \boldsymbol{j} \mid j_{\tau}=0 \land \gamma_{{n},\boldsymbol{j}}=0$ &
	$\blacktriangleright$ Remove unshipped job with due date 0 \\
	6:& \qquad $j_{\tau} \mapsfrom j_{\tau} - 1  \mid j_{\tau}>0$ & $\blacktriangleright$ Decrement time till due date\\
	7: & $\boldsymbol{J}_{{n}+1} \mapsfrom \boldsymbol{J}_{n}^x \cup \boldsymbol{\tilde{J}}_{{n}+1}$ &  $\blacktriangleright$ Merge existing and new job sets \\ 
	8: & Output: $\boldsymbol{J}_{{n}+1}$ & $\blacktriangleright$ New state\\ 
	\bottomrule  
\end{tabular}
\normalsize
\vspace{4mm}

\subsection{Policies and game-theoretical  properties}\label{ssec:problem_properties}
From a game-theoretical perspective, the system has some interesting properties that are worth exploring before turning to solutions. In infinitely repeated games without discounting (corresponding to our infinite horizon problem), a common objective is to maximize average profits \cite{friedman1971}. Following the conventions of the limit of means theorem, the optimal strategy (for the shipper, carrier is near-equivalent) look as follows:

\begin{align}\label{eq:optimal_policy_shp}
\pi^{S,*}=\lim_{n \mapsto \infty} \frac{1}{n} \sum_{n=0}^N   \sum_{\boldsymbol{j} \in \boldsymbol{J}_n}   \left(\sum_{n^{\prime}=n}^{n+|\mathcal{T}_{\boldsymbol{j}}|} \argmax_{b_{n^{\prime},\boldsymbol{j}} \in \mathbb{R}} \mathbb{E}(r_{\boldsymbol{j}}(\gamma_{n^{\prime},\boldsymbol{j}},b_{n^{\prime},\boldsymbol{j}}))\right)
\end{align}

The problem may be classified as a bargaining game \cite{nash1953}, in which the difference between transport costs and maximum willingness to pay is a surplus to be divided between shipper and carrier (remind the broker cannot actively influence the game). Both agents independently `claim' a proportion of the system-wide gain. The so-called feasibility set contains all solutions in which both agents achieve a nonnegative payoff. If no agreement is reached (i.e., when the cumulative claim exceeds the system-wide gain or agents can only earn non-positive payoffs), a disagreement point should exist. That is, each agent can execute a credible threat when the opponent deviates from the set of feasible solutions, as such capping its payoff. As shown in Lemma~2, each agent is able to cap the opponent's payoff at 0. We prove this result for the carrier; a similar proof may be constructed for the shipper.

\textbf{Lemma 2: Existence of disagreement point}\\
\textit{For the carrier, there exists a strategy $\pi^C$ that ensures the shipper's payoff $r_{\boldsymbol{j}}^S$ equals at most 0 for any job $\boldsymbol{j}$, regardless of the opponent's strategy $\pi^S$.}

\textbf{Proof:}
The full proof is found in Appendix A. If the carrier asks a price higher than the shipper's maximum willingness to pay, the shipper must either bid higher than that (resulting in a negative payoff when accepted) or bid below the ask and forfeit the agreement (payoff of 0). $\blacksquare$

Following Nash \cite{nash1953}, uncertainty vanishes in the limit and the game converges to an equilibrium. In such an equilibrium, agents cannot improve their payoff by unilaterally changing their strategy. The folk theorem \cite{friedman1971} states that an equilibrium payoff profile should satisfy two properties.  First, the payoff should be a convex combination of payoffs of the stage game, e.g., a weighted average as defined in Equation~\eqref{eq:optimal_policy_shp}.  Second, the equilibrium payoff must be individually rational, paying at least as much as the deviation point. Lemma~3 shows that the latter condition holds and formalizes the Nash equilibrium.
edible, because both players will never make any profit.

\textbf{Lemma 3: Existence of Nash equilibrium}
\textit{Any payoff profile satisfying $r_{\boldsymbol{j}}^C+r_{\boldsymbol{j}}^S \equiv c_{\boldsymbol{j}}^{S,max}-c_{\boldsymbol{j}}^{C,trn}$ is a Nash equilibrium. This payoff profile is achieved by any pair of strategies satisfying $c_{\boldsymbol{j}}^{C,trn}\leq a_{\boldsymbol{j}} = b_{\boldsymbol{j}} \leq c_{\boldsymbol{j}}^{S,max}$.}

\textbf{Proof:}
The full proof is found in Appendix A. Intuitively, when $a_{\boldsymbol{j}} < b_{\boldsymbol{j}}$, an agent could unilaterally improve its payoff without triggering the disagreement point. All profiles not satisfying $c_{\boldsymbol{j}}^{C,trn}\leq a_{\boldsymbol{j}} = b_{\boldsymbol{j}} \leq c_{\boldsymbol{j}}^{S,max}$ are outside the feasible set and invoke the disagreement point. $\blacksquare$

The key result of this section is that any payoff profile satisfying $c_{\boldsymbol{j}}^{C,trn} \leq a_{\boldsymbol{j}} = b_{\boldsymbol{j}} \leq c_{\boldsymbol{j}}^{S,max}$ is a Nash equilibrium. In Section~\ref{sec:experimental_design} we use this result to define appropriate measures for the experimental design.

\section{Solution method}\label{sec:solution_method}
This section explains the solution method, which is based on deep reinforcement learning techniques. Finding the optimal strategies to solve Equation~\eqref{eq:optimal_policy_shp} is hardly possible. The expectation depends on both the (unknown and potentially mixed) strategy of the opposing agent and the stochastic realization of new jobs; there is no guarantee that the optimal bid today will also succeed tomorrow. Furthermore, bid and ask prices are determined at the individual job level, whereas the broker allocates jobs at the state level. Finally, we deal with continuous action spaces that (naturally) contain infinitely many actions. For the above-mentioned reasons we resort to reinforcement learning to learn approximate strategies. Both shipper and carrier actively learn their pricing strategies based on observations. As changing the strategy of one agent influences the other, we are dealing with a highly non-stationary system.

In Section~\ref{ssec:policy_gradient} we present a policy gradient algorithm, in which the strategy is constantly adjusted to maximize expected rewards given the expected value. Section~\ref{ssec:extensions} presents several extensions to the base algorithm, including value function approximations (actor-critic models). Section~\ref{ssec:penalties} introduce a penalty structure, which is crucial to direct solutions towards the set of feasible solutions. Finally, Section~\ref{ssec:update_procedure} describes the update procedure.

\subsection{Policy gradient learning}\label{ssec:policy_gradient}
The reinforcement learning algorithms used in this paper are policy gradient algorithms. This class of algorithms directly operates on the strategy. Policy gradient methods can be expanded by adding a value function approximation to estimate downstream rewards related to current actions -- in that case we speak of actor-critic algorithms. We first discuss the vanilla policy gradient algorithm REINFORCE \cite{williams1992} and extend to actor-critic models in Section~\ref{ssec:extensions}. For readability, we only present notation for the shipper; carrier variants are near-identical.


In policy gradient reinforcement learning, actions are applied directly on the state. Every simulated episode $m \in \{0,1,\ldots,M\}$ yields a batch of job observations with corresponding rewards for all agents. The actions are determined by a stochastic strategy:

\begin{align}
 \pi_{\boldsymbol{\theta}^S}^{S,m}(b_{\boldsymbol{j}} \mid {\boldsymbol{j}},\boldsymbol{J}_{n})=\mathbb{P}^{\boldsymbol{\theta}^S}(b_{\boldsymbol{j}} \mid \boldsymbol{j},\boldsymbol{J}_{n})\enspace,
\end{align}

 where $\boldsymbol{\theta}^{S}$ is the parametrization of the strategy. The probability distributions representing the strategies are Gaussian distributions; bids (asks) are drawn independently for each container:
 
\begin{align}
 b_{\boldsymbol{j}} \sim \mathcal{N}(\mu_{\boldsymbol{\theta}^S}(\boldsymbol{j},\boldsymbol{J}_{n}), \sigma_{\boldsymbol{\theta}^S}), \forall \boldsymbol{j} \in \boldsymbol{J}_{n}\enspace.
 \end{align}
  
The parameterized standard deviation $\sigma_{\boldsymbol{\theta}^{S}}$ determines the level of exploration while learning, but is also an integral part of the strategy itself. Standard deviations typically decrease to small values once appropriate means are identified, but may also be larger when retaining some fluctuation is beneficial. For instance, strategies embedding randomness may be more difficult to counter.

The stored action-reward trajectories during each episode $m$ indicate which actions resulted in good rewards. We compute gradients which ensure that the strategy is updated in that direction. Only completed jobs (i.e., shipped or removed) can be used to update the strategy; we introduce some additional notation for this purpose. Let $\boldsymbol{K}^m=[K_0^m,\ldots,K_{\tau^{max}}^m]$ be a vector containing the number of observed bid-ask pairs for completed jobs. For instance, if a job had an original due date of 3 and is shipped at $j_{\tau}=1$, we  increment $K_{3}^m$, $K_{2}^m$ and $K_{1}^m$ by 1, using an update function $k(\boldsymbol{j})$. Completed jobs are stored in a set $\hat{\boldsymbol{J}}^m$. For each episode the cumulative rewards per job -- shown for the shipper here -- are defined as follows:
\begin{align}
\hat{v}_{t,\boldsymbol{j}}^{S,m}(\gamma_{t,\boldsymbol{j}},b_{t,\boldsymbol{j}})= 
\begin{cases}
r_{t,\boldsymbol{j}}(\gamma_{t,\boldsymbol{j}},b_{t,\boldsymbol{j}})^{S,m} &  \text{if} \qquad t=0 \\
r_{t,\boldsymbol{j}}(\gamma_{t,\boldsymbol{j}},b_{t,\boldsymbol{j}})^{S,m}+\hat{v}_{t-1,\boldsymbol{j}}^{S,m} & \text{if} \qquad t>0 \notag
\end{cases}
\quad, \forall t \in \mathcal{T}_{\boldsymbol{j}}\enspace.
\end{align}


The cumulative rewards observed at time $n$ in episode $m$ are stored in vectors $\hat{\boldsymbol{v}}_{n}^{S,m}=\bigl[[\hat{v}_{t,\boldsymbol{j}}^{S,m}]_{t \in \mathcal{T}_{\boldsymbol{j}}}\bigr]_{\forall \boldsymbol{j} \in \boldsymbol{J}_t}$ and $\hat{\boldsymbol{v}}_{n}^{C,m}=\bigl[[\hat{v}_{t,\boldsymbol{j}}^{C,m}]_{t \in \mathcal{T}_{\boldsymbol{j}}}\bigr]_{\forall \boldsymbol{j} \in \boldsymbol{J}_t}$. At the end of each episode, we can construct the information vector:

\begin{align}
\boldsymbol{I}^{S,m} =\biggl[[\boldsymbol{J}_{n}^m, \boldsymbol{b}_{n}^m, \hat{\boldsymbol{v}}_{n}^{S,m},
\boldsymbol{\gamma}_{n}^m]_{\forall n \in \{0,\ldots,N\}}, 
\boldsymbol{K}^m, \hat{\boldsymbol{J}}^m\biggr] \notag \enspace.
\end{align}

\noindent The information vector contains the states, actions and rewards  required for the strategy updates. For this purpose we utilize the policy gradient theorem; see Sutton \& Barto \cite{sutton2018} for a detailed description. We present the theorem for the shipper:
\begin{align}	
\nabla_{\boldsymbol{\theta}^S} v_{j_\tau,\boldsymbol{j}}^{\pi_{\boldsymbol{\theta}^S}}
\propto \sum_{{n}=1}^{N} \left(\int_{\boldsymbol{J}_{n} \in \mathcal{J}_{n}} \mathbb{P}^{\pi_{\boldsymbol{\theta}}^S}(\boldsymbol{J}_{n} \mid \boldsymbol{J}_{{n}-1}) \int_{b_{\boldsymbol{j}} \in \mathbb{R}}   \nabla_{\boldsymbol{\theta}^S}{\pi_{\boldsymbol{\theta}}^S}(b_{\boldsymbol{j}} \mid \boldsymbol{j}, \boldsymbol{J}_{n})v_{j_\tau,\boldsymbol{j}}^{\pi_{\boldsymbol{\theta}}^S}(\gamma_{\boldsymbol{j}},b_{\boldsymbol{j}})\right)  \enspace.  \notag
\end{align}
We proceed to apply the policy gradient theorem to our system, adopting a Gaussian decision-making strategy. We use a neural network (actor network) to output the Gaussian distribution. Let $\boldsymbol{\theta}^S$ define the set of weight parameters describing the decision-making strategy $\pi_{\boldsymbol{\theta}^S} \colon (\boldsymbol{j},\boldsymbol{J}_{n}) \mapsto b_{\boldsymbol{j}}$. Furthermore, let $\boldsymbol{\phi}(\boldsymbol{j},\boldsymbol{J}_{n})$ be a feature vector distilling the most salient state attributes, for instance the average time till due date or the number of jobs waiting. The features used for our study are described in Section~\ref{ssec:features}. For the actor network, the feature vector $\boldsymbol{\phi}$ is the input,  $\boldsymbol{\theta}^S$ represents the network weights, and the mean bid (ask) $\mu_{\boldsymbol{\theta}^S}$ and standard deviation $\sigma_{\boldsymbol{\theta}^S}$ are the output. We formalize the strategy as follows \footnote{$\pi_{\boldsymbol{\theta}^S}$ is the parameterized strategy, $\pi$ represents the mathematical constant}:
\begin{align}
\pi_{\boldsymbol{\theta}^S}=\frac{1}{\sqrt{2\pi}\sigma_{\boldsymbol{\theta}}}e^{-\frac{\left(b_{\boldsymbol{j}}-\mu_{\boldsymbol{\theta}^S}\left(\boldsymbol{j},\boldsymbol{J}_{n}\right)\right)^2}{2\sigma_{\boldsymbol{\theta}^S}^2}}\enspace. \notag
\end{align}
with $b_{\boldsymbol{j}}$ being the bid price, $\mu_{\boldsymbol{\theta}^S}(\boldsymbol{j},\boldsymbol{J}_{n})$ the Gaussian mean and $\sigma_{\boldsymbol{\theta}^S}$ the parametrized standard deviation. The corresponding action $b_{\boldsymbol{j}}$ is acquired from the inverse normal distribution. Parameter updates take place after each episode, utilizing a function $ U({\boldsymbol{\theta}}^{S,m},\boldsymbol{I}^{S,m})$. We describe the update procedure in more detail in Section~\ref{ssec:update_procedure}.

The core concept behind the policy gradient algorithm is that the normal distribution constituting the strategy converges to a distribution of prices appropriate for the state. Actions with high rewards and low probabilities yield the strongest update signals. The algorithmic outline to update the parametrized strategy is outlined in Algorithm~2.

\setcounter{algo}{1}
\begin{algo}\label{policygradientoutline}Outline of the policy gradient  algorithm (based on \cite{williams1992})
\end{algo}
\small
\begin{tabular}{ l l l}
	\toprule		
	0: & Input:  $\pi_{\boldsymbol{\theta}^{S}}^0,\pi_{\boldsymbol{\theta}^{C}}^0$ & $\blacktriangleright$ Differentiable parametrized strategies\\	
	1: & $\boldsymbol{\theta}^{S} \mapsfrom \mathbb{R}^{|\boldsymbol{\theta}^{S}|},\boldsymbol{\theta}^{C} \mapsfrom \mathbb{R}^{|\boldsymbol{\theta}^{C}|}$ & $\blacktriangleright$ Initialize parameters\\
	2: & $\forall m \in \{0,\ldots,M\}$ & $\blacktriangleright$ Loop over episodes\\
	3: & \qquad $\boldsymbol{\hat{J}}^m \mapsfrom \emptyset$ & $\blacktriangleright$ Initialize completed job set\\
	4: & \qquad $\boldsymbol{J}_0 \mapsfrom \mathcal{J}_0$ & $\blacktriangleright$ Generate initial state\\
	5: & \qquad $\forall n \in \{0,\ldots,N\}$ & $\blacktriangleright$ Loop over finite time horizon\\
	6: & \qquad\qquad  $b_{\boldsymbol{j}}^m \mapsfrom \pi_{\boldsymbol{\theta}^S}^m(\boldsymbol{j},\boldsymbol{J}_{n}), \forall \boldsymbol{j} \in \boldsymbol{J}_{n}$ & $\blacktriangleright$ Bid placement jobs\\
	7: & \qquad\qquad  $a_{\boldsymbol{j}}^m \mapsfrom \pi_{\boldsymbol{\theta}^C}^m(\boldsymbol{j},\boldsymbol{J}_{n}), \forall \boldsymbol{j} \in \boldsymbol{J}_{n}$ & $\blacktriangleright$ Ask placement jobs\\
	8: & \qquad\qquad  $\boldsymbol{\gamma}^m \mapsfrom \argmax_{\boldsymbol{\gamma}^m \in \Gamma(\boldsymbol{J}_{n})} $ & $\blacktriangleright$ Job allocation broker, Eq. \eqref{eq:selectionbroker}\\
	& \qquad\qquad $\left(\sum_{\boldsymbol{j} \in \boldsymbol{J}_{t^\prime}} {\gamma}_{\boldsymbol{j}}^m(b_{\boldsymbol{j}} - a_{\boldsymbol{j}})\right)$ & \\
	9a: & \qquad\qquad $\hat{v}_{j_\tau,\boldsymbol{j}}^{S,m} \mapsfrom r_{\boldsymbol{j}}(\gamma_{\boldsymbol{j}}^m,b_{\boldsymbol{j}}^m), \forall \boldsymbol{j} \in \boldsymbol{J}_{n}$ & $\blacktriangleright$ Compute cumulative rewards shipper\\ 
	9b: & \qquad\qquad $\hat{v}_{j_\tau,\boldsymbol{j}}^{C,m} \mapsfrom r_{\boldsymbol{j}}(\gamma_{\boldsymbol{j}}^m,b_{\boldsymbol{j}}^m), \forall \boldsymbol{j} \in \boldsymbol{J}_{n}$ & $\blacktriangleright$ Compute cumulative rewards carrier\\ 
	10: & \qquad\qquad $\forall \boldsymbol{j} \in \boldsymbol{J}_{n} \mid j_{\tau}=0 \lor {\gamma}_{\boldsymbol{j}}^m=1 $& $\blacktriangleright$ Loop over completed jobs \\
	11: & \qquad\qquad\qquad $\boldsymbol{\hat{J}}^m \mapsfrom \boldsymbol{\hat{J}}^m \cup \{\boldsymbol{j}\}$ & $\blacktriangleright$ Update set of completed jobs\\ 
	12: & \qquad\qquad\qquad $\boldsymbol{K}^m \mapsfrom k(\boldsymbol{j})$ & $\blacktriangleright$ Update number of completed jobs\\
	13: & \qquad\qquad $\boldsymbol{\tilde{J}}_{n}\mapsfrom \mathcal{\tilde{J}}_{n}$ & $\blacktriangleright$ Generate job arrivals\\
	14: & \qquad\qquad $\boldsymbol{J}_{n+1} \mapsfrom S^M(\boldsymbol{J}_{n},\boldsymbol{\tilde{J}}_{n+1},\boldsymbol{\gamma}_{t^\prime}^m) $ & $\blacktriangleright$ Transition function, Algorithm 1\\
	15a: & \qquad $\boldsymbol{I}^{S,m} \mapsfrom \biggl[[\boldsymbol{J}_{n}^m, \boldsymbol{b}_{n}^m, \hat{\boldsymbol{v}}_{n}^{S,m},
	\boldsymbol{\gamma}_{n}^m]_{\forall n \in \{0,\ldots,N\}}, 
	\boldsymbol{K}^m, \hat{\boldsymbol{J}}^m\biggr]$ & $\blacktriangleright$ Store information shipper\\
	15b: & \qquad $\boldsymbol{I}^{C,m} \mapsfrom \biggl[[\boldsymbol{J}_{n}^m, \boldsymbol{a}_{n}^m, \hat{\boldsymbol{v}}_{n}^{C,m},
	\boldsymbol{\gamma}_{n}^m]_{\forall n \in \{0,\ldots,N\}}, 
	\boldsymbol{K}^m, \hat{\boldsymbol{J}}^m\biggr]$ & $\blacktriangleright$ Store information carrier\\
	16a:& \qquad ${\boldsymbol{\theta}}^{S,m+1} \mapsfrom U({\boldsymbol{\theta}}^{S,m},\boldsymbol{I}^{S,m}) $ & $\blacktriangleright$ Update actor network shipper \\
	16b:& \qquad ${\boldsymbol{\theta}}^{C,m+1} \mapsfrom U({\boldsymbol{\theta}}^{C,m},\boldsymbol{I}^{C,m}) $ & $\blacktriangleright$ Update actor network carrier \\
	17: & Output: $\pi_{\boldsymbol{\theta}^S}^M,\pi_{\boldsymbol{\theta}^C}^M$ & $\blacktriangleright$ Return tuned strategies\\
	\bottomrule  
\end{tabular}
\normalsize

\subsection{Policy gradient extensions}\label{ssec:extensions}
In Section~\ref{ssec:policy_gradient} we presented the basic policy gradient algorithm. We now introduce four extensions that will be tested, namely (i) policy gradient with baseline, (ii) actor-critic with Q-value, (iii) temporal difference learning -- also known as TD(1) -- and (iv) actor-critic with advantage function. The algorithms are summarized in Table~\ref{table:action_value}; for detailed descriptions we refer to  Sutton \& Barto \cite{sutton2018}.

\begin{table}[h!]
	\scriptsize
	\caption{Algorithmic variants of the policy gradient algorithmic.}
	\label{table:action_value}       
	\begin{tabular}{lll}
		\hline\noalign{\smallskip}
		Algorithm & Reward signal & Description  \\
		\hline\noalign{\smallskip}
		Policy gradient & $\hat{v}_{n,\boldsymbol{j}}^m$ & Cumulative reward  \\
		Policy gradient with baseline &  $\hat{v}_{n,\boldsymbol{j}}^m-\bar{v}_{j_\tau}^m$ & Deduct average reward \\
		Q-value & $Q(b_{\boldsymbol{j}},\boldsymbol{j},\boldsymbol{J}_{n})$ & Estimated cumulative reward \\
		TD(1) & $\hat{v}_{n,\boldsymbol{j}}^{m}-Q(b_{\boldsymbol{j}},\boldsymbol{j},\boldsymbol{J}_{n})$ & Deduct Q-value \\
		Advantage value & $Q(b_{\boldsymbol{j}},\boldsymbol{j},\boldsymbol{J}_{n})-Q(\mu_{\boldsymbol{\theta}},\boldsymbol{j},\boldsymbol{J}_{n})$ & Deduct average Q-value  \\
		\noalign{\smallskip}\hline
	\end{tabular}
\end{table}

We first discuss the policy gradient with baseline. Rewards may exhibit large variance that hampers learning. To reduce this variance, we deduct the average observed value $\bar{v}_{j_\tau}^m$ during the episode as a baseline value \cite{sutton2018}. In our update procedure, we then replace $\hat{v}_{n,\boldsymbol{j}}^m$ with $\hat{v}_{n,\boldsymbol{j}}^m-\bar{v}_{j_\tau}^m$, yielding lower variance than the original. For the prevailing episode $m$, the baseline is defined by 
\begin{align}
\bar{v}_{j_\tau}^m = \frac{1}{K_{j_\tau}^m} \sum_{\boldsymbol{j} \in \boldsymbol{\hat{J}}^m} \hat{v}_{n,\boldsymbol{j}}^m, \forall {j_\tau} \in \{0,\ldots, \tau^{max}\}  \enspace. \notag
\end{align}

\begin{figure}
	\includegraphics[width=0.5\textwidth]{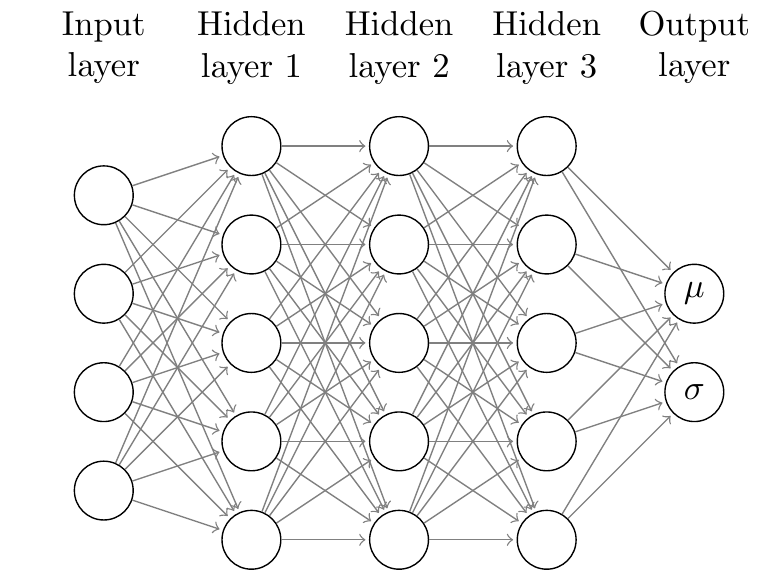}%
	\includegraphics[width=0.5\textwidth]{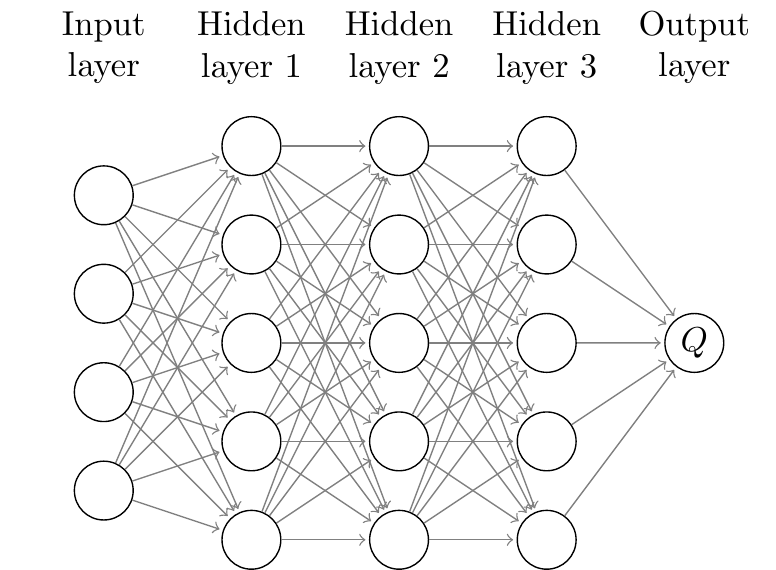}%
	\caption{Actor network (left) outputting $\mu$ and $\sigma$, critic network (right) outputting $\bar{V}$.}%
	\label{fig:actor_critic_networks}
\end{figure}

The second extension is the actor-critic algorithm with Q-values, which may be seen as a hybrid between policy approximation and value approximation. Figure~\ref{fig:actor_critic_networks} provides an illustration of an actor-critic architecture. Policy gradient methods rely on directly observed rewards, which may strongly vary between episodes. Furthermore, actor networks do not leverage information about particular states encountered. In the actor-critic approach, we replace the observed value $\hat{v}$ with a  function $Q(b_{\boldsymbol{j}},\boldsymbol{j},\boldsymbol{J}_{n})$ that is often defined by a neural network (critic network). The critic network transforms the input features into an expected reward value, popularly known as a Q-value. Drawbacks of actor-critic methods are (i) the need to learn additional parameters, (ii) slower convergence than actor-only methods and (iii) simultaneous adjustments of strategy and value function. Particularly the latter is problematic in highly non-stationary multi-agent settings; value functions learned in the past may no longer be representative for current strategies and vice versa.


 
Third, temporal difference learning -- denoted by TD(1) -- or Monte Carlo reinforcement learning utilizes the Q-value as a baseline by subtracting it from the observed rewards, yielding a reward signal $\hat{v}_{n,\boldsymbol{j}}^{m}-Q(b_{\boldsymbol{j}},\boldsymbol{j},\boldsymbol{J}_{n})$. This approach preserves both the actual observations and the value function approximations while reducing variance.

The fourth extension we discuss is the advantage function (also known as Advantage Actor Critic or A2C). It also uses a baseline function, but utilizes value functions rather than reward observations. Specifically, we define the reward signal
$Q(b_{\boldsymbol{j}},\boldsymbol{j},\boldsymbol{J}_{n})-Q(\mu_{\boldsymbol{\theta}},\boldsymbol{j},\boldsymbol{J}_{n})$, where the baseline term is a reward function that depends on the state but is independent of the action (bid or ask). Concretely, the first Q-value uses the sampled bid/ask as a feature, the second Q-value the mean bid/ask.  Again, the objectives are to reduce the variance and to generalize past observations.



\subsection{Penalties}\label{ssec:penalties}
The reward functions presented in Section~\ref{ssec:decisions} return 0 in case of a failed job. Such a reward signal yields no update direction towards profit-generating solutions and may trap the algorithm in a poor local optimum. To resolve this problem, we add variable penalties to nudge the behavior of agents. Penalties are a function of the bid- or ask price that provides direction. The lower the bid (higher the ask), the stronger the penalty when it is rejected. For the shipper, we have

 \begin{align}\label{eq:rewardshipperpenalty}
 r_{\boldsymbol{j}}^{S}(\gamma_{\boldsymbol{j}},b_{\boldsymbol{j}})= 
 \begin{cases}
 c_{\boldsymbol{j}}^{S,max}- b_{\boldsymbol{j}} &  \text{if} \qquad \gamma_{\boldsymbol{j}}=1  \\
\max(0,c_{\boldsymbol{j}}^{S,max}-b_{\boldsymbol{j}}) & \text{if} \qquad \gamma_{\boldsymbol{j}}=0 \\
 \end{cases}\enspace.
 \end{align}
 
For the carrier, incurring a penalty only makes sense when (i) the ask price exceeded the shipment costs and (ii) there was sufficient idle capacity to accommodate the rejected job. Thus, we modify the reward function as follows:

\begin{align}\label{eq:rewardcarrierpenalty}
r_{\boldsymbol{j}}^{C}(\gamma_{\boldsymbol{j}},a_{\boldsymbol{j}})= 
\begin{cases}
a_{\boldsymbol{j}}- c_{\boldsymbol{j}}^{C,trn} &  \text{if} \qquad \gamma_{\boldsymbol{j}}=1 \\
\max(0,a_{\boldsymbol{j}}- c_{\boldsymbol{j}}^{C,trn}) & \text{if} \qquad \gamma_{\boldsymbol{j}}=0 \land
 \zeta^{C} - \left(\sum_{\boldsymbol{j} \in \boldsymbol{J}_{n}} {\gamma}_{\boldsymbol{j}} \cdot j_v \right) >0\\
0 & \text{otherwise} \\
\end{cases}\enspace.
\end{align}



\subsection{Update procedure}\label{ssec:update_procedure}

Algorithm~2 outlined the generic policy gradient algorithm with a generic update function $ U({\boldsymbol{\theta}}^{S,m},\boldsymbol{I}^{S,m})$. Here we treat the updating procedure in more detail. Traditionally, policy gradient algorithms follow stochastic gradient ascent for updates. The corresponding gradients can be computed with respect to each feature and are defined by
\begin{align}
\nabla_{\mu_{\boldsymbol{\theta}}}(\boldsymbol{j},\boldsymbol{I}^m) & \frac{(b_{\boldsymbol{j}}-\mu_{\boldsymbol{\theta}}(\boldsymbol{j},\boldsymbol{J}_{n}))\phi(\boldsymbol{j},\boldsymbol{J}_{n})}{\sigma_{\boldsymbol{\theta}}^2} \enspace, \notag\\
\nabla_{\sigma_{\boldsymbol{\theta}}} (\boldsymbol{j},\boldsymbol{I}^m)  &= \frac{(b_{\boldsymbol{j}}-\mu_{\boldsymbol{\theta}}(\boldsymbol{j},\boldsymbol{J}_{n}))^2 - \sigma_{\boldsymbol{\theta}}^2}{\sigma_{\boldsymbol{\theta}}^3}\enspace. \notag
\end{align}

In a neural network setting, we might compute these gradients with respect to the activation functions and determine the corresponding updates for the network weights (classical stochastic gradient ascent). However, it is often convenient to define a loss function that allows updating with state-of-the-art gradient \textit{descent} algorithms, using a backpropagation procedure. Update algorithms such as ADAM often outperform traditional gradient descent. The Gaussian loss function \cite{vanheeswijk2019b} is defined by:

\begin{equation}\label{eq:gaussianlossfunction}
\mathcal{L}^{actor}(b_{\boldsymbol{j}},\boldsymbol{J}_{n},\hat{v}_{\boldsymbol{j}}) = - \log\left(\frac{1}{\sqrt{2\pi}\sigma_{\boldsymbol{\theta}}}e^{-\frac{\left(b_{\boldsymbol{j}}-\mu_{\boldsymbol{\theta}}\left(\boldsymbol{j},\boldsymbol{J}_{n}\right)\right)^2}{2\sigma_{\boldsymbol{\theta}^S}^2}}\right) \hat{v}_{\boldsymbol{j}}\enspace.
\end{equation}

To update critic networks, we also start with a loss function such that we can perform backpropagation. This loss function is simply the mean-squared error between Q-value and observed rewards:

\begin{align}
\mathcal{L}^{critic}(b_{\boldsymbol{j}},\boldsymbol{J}_{n},\hat{v}_{\boldsymbol{j}})=(\hat{v}_{j_\tau,\boldsymbol{j}}^{m}-Q(b_{\boldsymbol{j}},\boldsymbol{j},\boldsymbol{J}_{n}))^2\enspace.
\end{align}

\section{Experimental design}\label{sec:experimental_design}
This section introduces the experimental design. The main objective is to provide insights into the behavior of the algorithm. Section~\ref{ssec:case_properties} introduces the two test cases and Section~\ref{ssec:performance_metrics} defines the performance metrics. Section~\ref{ssec:implementation} describes the Python implementation. Finally, Section~\ref{ssec:features} lists the features used as network input.

\subsection{Case properties}\label{ssec:case_properties}
We present two test cases for this study; a toy-sized deterministic one (Case I) and a larger stochastic one (Case II). 

In Case I (deterministic), each day exactly one job arrives with due date $j_\tau=0$, volume $j_v=\zeta^C$ and a fixed distance $j_d$. It follows that the job is shipped if $a_j<b_j$ and fails otherwise. The maximum willingness to pay is 2, the transport costs are 1. This simple setting allows to test behavior in-depth under a variety of circumstances. We use this case to tune parameters, explore the parameter space and obtain behavioral insights.

Case II stochastically generates a number of job arrivals each day and includes varying job properties. Due dates, volumes and distances range between 1 and 5. Per volume unit per mile, the maximum willingness to pay is 2 and transport costs are 1. The number of jobs arriving daily varies between 0 and 10; the total number of jobs may accumulate up to 50. We consider two variants of the case, one in which the transport capacity is 40 (somewhat scarce) and one where it is 300 (abundant). The additional challenges in Case II, compared to Case I, stem from the uncertain availability of sufficient capacity and the varying dimensions of the jobs. Table~\ref{table:case settings} summarizes the case settings.

\begin{table}
	\scriptsize
	\centering
	\caption{Settings for Case I (deterministic) and Case II (stochastic).}
	\label{table:case settings}
	\begin{tabular}{ l l  l }
		\toprule		
		Setting & Case I & Case II\\
		\midrule
		Number of job arrivals & 1& [1-10] \\
		Due date & 0 & [1-5] \\
		Job transport distance & 1& [1-5] \\
		Job volume &1 & [1-5]\\
		Willingness to pay (per volume unit per mile)&2 & 2\\
		Transport costs (per volume unit per mile) &1& 1 \\
		Transport capacity &1 & $\{40, 300\}$ \\
		\bottomrule  
	\end{tabular}
\end{table}

\subsection{Performance metrics}\label{ssec:performance_metrics}
The objective of our design is to represent a completely decentralized, self-organizing transport market without interventions or regulations. The performance metrics are designed in line with this purpose. We study an online environment, in which strategies are continuously updated while measuring performance.

In Section~\ref{ssec:problem_properties}, we established that infinitely many Nash equilibria exist for our system. However, the Nash equilibrium is a theoretical result that emerges in the limit after all uncertainty has been resolved. The setting studied here is inherently uncertain. Therefore, adherence to the Nash equilibrium is the first performance metric. Formally, we define adherence as follows:

\begin{align}
\max\left(0,\sum\limits_{t \in \mathcal{T}_{\boldsymbol{j}}} \gamma_{t,{\boldsymbol{j}}}
\frac{(a_{\boldsymbol{j}}-c_{\boldsymbol{j}}^{min}) +(c_{\boldsymbol{j}}^{max} - b_{\boldsymbol{j}})}{(c_{\boldsymbol{j}}^{max}-c_{\boldsymbol{j}}^{min})}\right)\notag \enspace.
\end{align}


The second performance metric is fairness. Although a Nash equilibrium might be rational, it is not necessarily perceived as fair; one agent might receive 0 and the other $c^{S,max}-c^{C,trn}$. In a completely fair setting both agents divide profits equally, in an unfair setting one agent reaps most of the rewards. We note that agents compete via their algorithms to increase their own reward share. As such, unfairness is an inherent aspect of competitive markets and not necessarily an indication of poor performance. We formalize fairness with the following metric:

\begin{align}
\max\left(0,1-\left|\sum\limits_{t \in \mathcal{T}_{\boldsymbol{j}}} \gamma_{t,{\boldsymbol{j}}} \frac{(a_{\boldsymbol{j}}-c_{\boldsymbol{j}}^{min}) -(c_{\boldsymbol{j}}^{max} - b_{\boldsymbol{j}})}{(a_{\boldsymbol{j}}-c_{\boldsymbol{j}}^{min}) +(c_{\boldsymbol{j}}^{max} - b_{\boldsymbol{j}})}\right|\right)\notag \enspace.
\end{align}


The third metric is utilization. If a transport service departs with idle capacity while feasible jobs (in terms of capacity) remain unshipped, this indicates an inefficiency in the market. Also note that the preceding metrics suffer with low utilization. To determine the upper bound of utilized capacity, we solve a variant of the knapsack problem in which we substitute the job value with the job volume, thus maximizing utilization of transport capacity:

\begin{align}\label{eq:selectionbrokervolume}
\argmax_{\boldsymbol{\gamma}_{\boldsymbol{j}}^{\zeta^C}\in \Gamma^{\zeta^C}(\boldsymbol{J}_{n})} \left(\sum_{\boldsymbol{j} \in \boldsymbol{J}_{n}} {\gamma}_{\boldsymbol{j}}^{\zeta^C} \cdot j_v\right)\enspace,
\end{align}
s.t.
\begin{align}
\sum_{\boldsymbol{j} \in \boldsymbol{J}_{n}} {\gamma}_{\boldsymbol{j}}^{\zeta^C} \cdot j_v \leq \zeta^C\enspace. \notag
\end{align}

Dividing the true utilization by the theoretically possible utilization gives the relative utilization:

\begin{align}
\frac{\sum_{\boldsymbol{j} \in \boldsymbol{J}_{n}} {\gamma}_{\boldsymbol{j}} \cdot j_v}
{\sum_{\boldsymbol{j} \in \boldsymbol{J}_{n}} {\gamma}_{\boldsymbol{j}}^{\zeta^C} \cdot j_v}
\end{align}








\subsection{Implementation}\label{ssec:implementation}
The algorithm is coded in Python 3.8 and uses TensorFlow 2.3 for the neural network architectures. To initialize the neural network weights, we use He initialization \cite{he2015} for all but the last layer. For the final layer of the actor networks, we initialize bias weights such that the carrier's ask equals $c^{C,trn}$ and the shipper's bid equals $c^{S,max}$ (we test different bias initializations in later experiments). For the Q-value (critic network), all weights in the final layer are initialized at 0, guaranteeing an unbiased initial estimate that does not influence decisions in the early iterations. We use the ADAM optimizer for network updates, which we find to consistently outperform optimizers such as classic stochastic gradient descent and RMSprop.

All solution methods are verified by checking the behavior of trivial strategies (e.g., by fixing the strategy of the opposing agent). We remark that our actor-critic methods (requiring to learn a Q-value) generally take long to converge even in simple settings, but their eventual convergence is confirmed.

We perform 5 replications for the main experiments (after the initial explorations), which in stable experiments gives standard deviations less than 5\% on the KPIs. We report average values (excluding the first 10\% of iterations, which serve as a warm-up period) and end-of-horizon results. The latter provide insights regarding convergence.

\subsection{Features}\label{ssec:features}
To parametrize the strategy we use several features, denoted by vector $\boldsymbol{\phi}$. We adopt the same set of features as defined in Van Heeswijk \cite{vanheeswijk2020b} for the single-agent system. For computing Q-values, we add the selected bid- or ask price to the set; for the advantage function we use the average bid/ask price. Table~\ref{table:features} lists the set of features. We note that by initializing weights corresponding to the bias, we can determine the initial values for $\mu_{\boldsymbol{\theta}}^0$ and $\sigma_{\boldsymbol{\theta}}^0$. In the implementation, all features are normalized before being used as input for the neural networks.

\begin{table}
	\scriptsize
	\centering
	\caption{Features describing the key state attributes, used as input for the actor- and critic networks.}
	\label{table:features}
	\begin{tabular}{ l  l }
		\toprule
		Feature & Description\\	
		\midrule	
		1 & Bias \\
		$j_\tau$ & Job due date \\
		$j_d$  & Job transport distance \\
		$j_v$  &Job volume\\
		$\frac{\sum_{\boldsymbol{j}\in \boldsymbol{J}_{n}}j_\tau}{|\boldsymbol{J}_{n}|}$ & Average due date\\
		$\frac{\sum_{\boldsymbol{j}\in \boldsymbol{J}_{n}}j_d}{|\boldsymbol{J}_{n}|}$ & Average distance\\
		$\frac{\sum_{\boldsymbol{j}\in \boldsymbol{J}_{n}}j_v}{|\boldsymbol{J}_{n}|}$ & Average volume \\
		$\sum_{\boldsymbol{j}\in \boldsymbol{J}_{n}}j_v$ & Total volume \\
		$|\boldsymbol{J}_{n}|$ & Total number of jobs \\
		$a_{\boldsymbol{j}}$ & Ask (only for Q-value carrier) \\
		$b_{\boldsymbol{j}}$ & Bid (only for Q-value shipper) \\
		\bottomrule  
	\end{tabular}
\end{table}


\section{Numerical results and analysis}\label{sec:numerical_results}

This section presents the results and analysis of the numerical experiments, as outlined in Section~\ref{sec:experimental_design}. In Section~\ref{ssec:preliminary} we explore the parameter space and determine suitable hyper-parameters. Section~\ref{ssec:analysis_case1}  analyzes behavior for the deterministic Case I, Section~\ref{ssec:analysis_case2} for the stochastic Case II.

\subsection{Exploration of parametric space}\label{ssec:preliminary} 
The experiments in this section (utilizing Case I) serve to set suitable hyper-parameters and to provide insights into various architectural settings.  The aspects we test are (i) number and length of episodes, (ii) learning rates, (iii) initial standard deviations, (iv) actor network architectures, (v) actor-critic variants and (vi) critic network architectures. In the result tables, we provide average results (excluding the 10\% warm-up period) and end-of-horizon results between parentheses.
\subsubsection{Verification}
\begin{figure}[ht] 
	\begin{minipage}{.5\textwidth}
		\includegraphics[width=\textwidth]{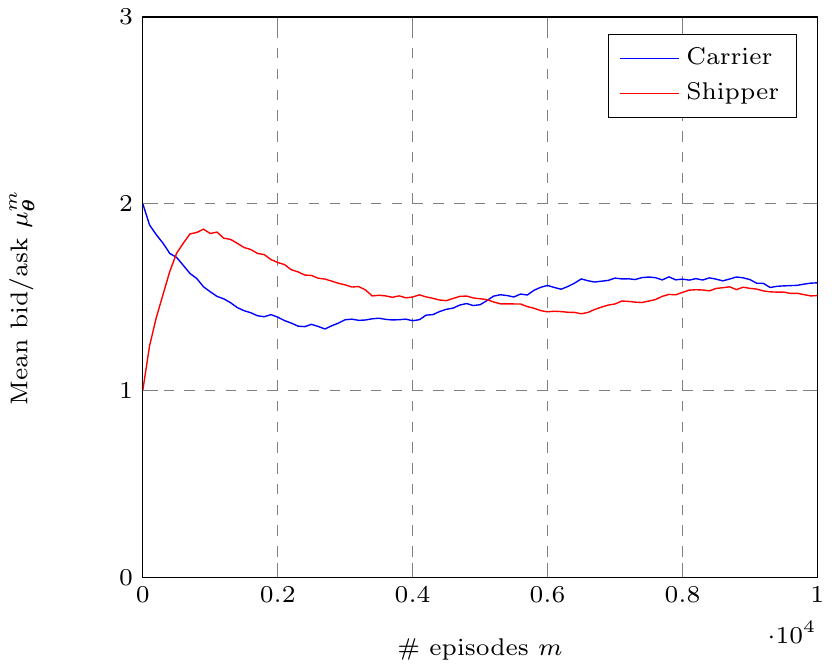}
		\caption{Conv. with transport costs 1 and maximum willingness 2}
		\label{fig:1a}       
	\end{minipage}
	\begin{minipage}{.5\textwidth}
		\includegraphics[width=\textwidth]{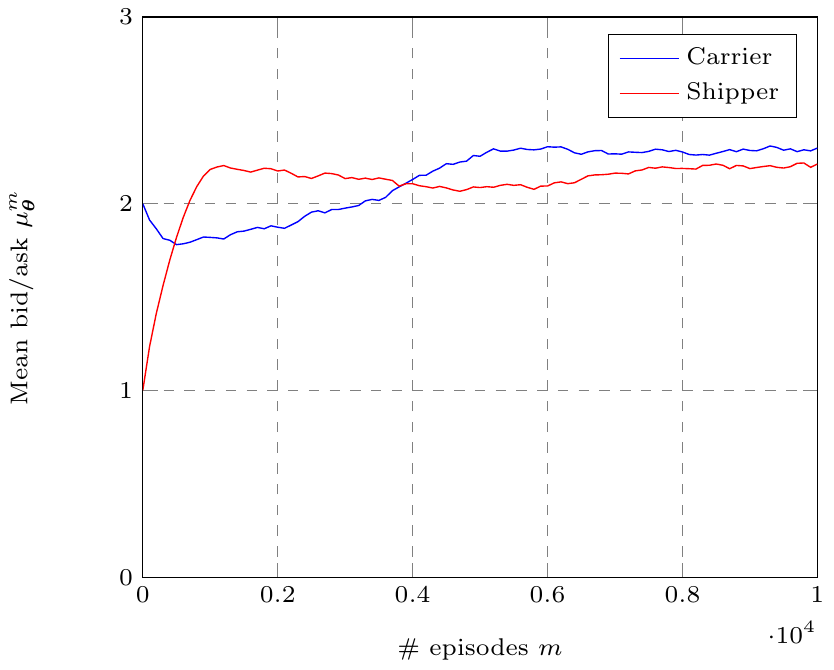}
		\caption{Conv. with transport costs 1.5 and maximum willingness 3}
		\label{fig:1b}       
	\end{minipage}
	\begin{minipage}{.5\textwidth}
		\includegraphics[width=\textwidth]{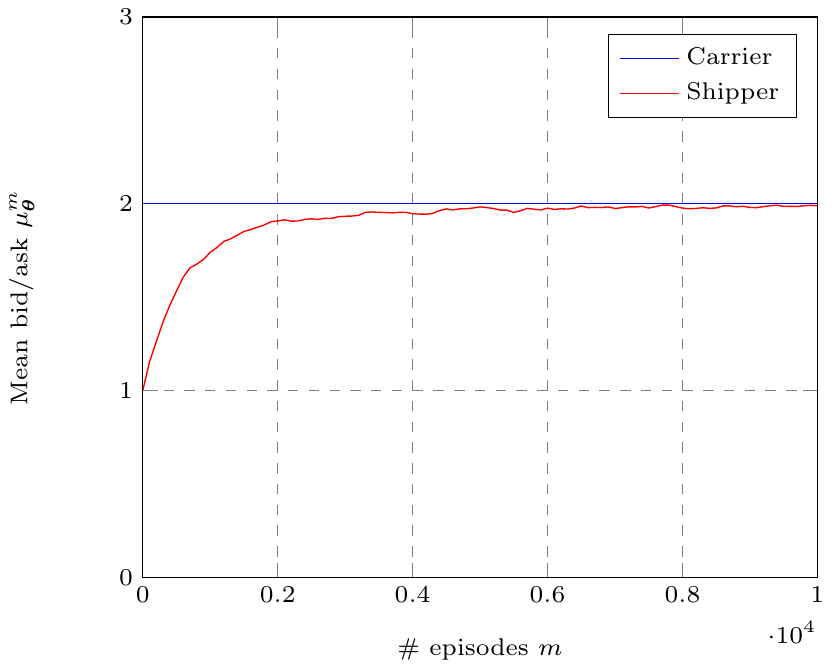}
		\caption{Conv. with bid fixed at 2.0}
		\label{fig:1c}       
	\end{minipage}
	\begin{minipage}{.5\textwidth}
		\includegraphics[width=\textwidth]{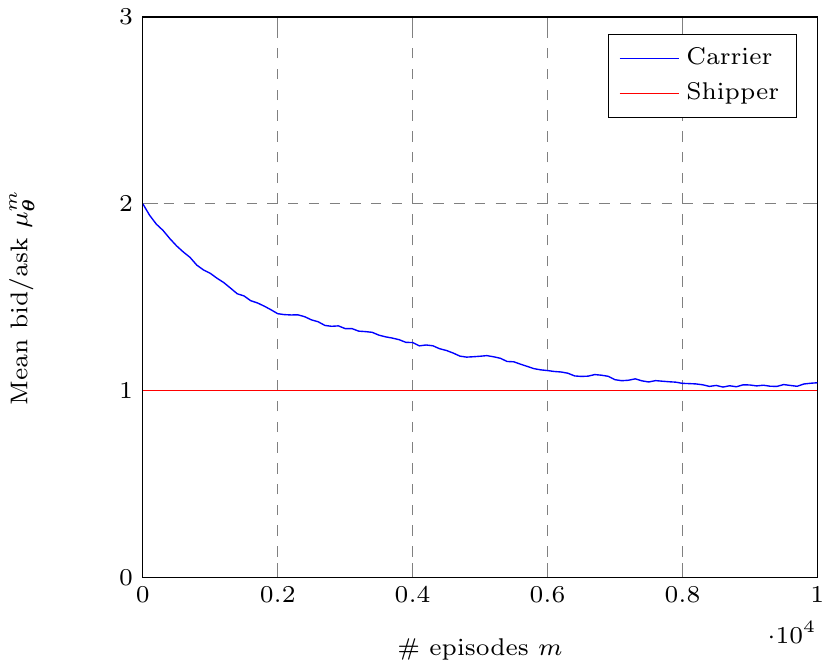}
		\caption{Conv. with ask fixed at 1.0}
		\label{fig:1d}       
	\end{minipage}
\end{figure}

We highlight a few verification experiments to demonstrate the behavior of the algorithm, using simplified system settings. If we fix a single agent's bid/ask causes the other agent's bid/ask to converge to near the fixed value, as long as it is within the feasible region. If both agent adopt equivalent strategies, we find that bid- and ask prices gravitate roughly towards the average of maximum willingness and transport costs, i.e., a fair balance. Figures~\ref{fig:1a}-\ref{fig:1d} illustrates some of the initial (pre-finetuning) behaviors observed during verification. 

\subsubsection{Episode lengths}
We test the effects of adjusting the number and length of episodes,  Smaller batches of observations result in more frequent strategy updates, yet these updates may be less stable. We keep the total number of simulated days fixed at 1 million, selecting $m \in \{10 ,100, 1000, 10000\}$ episodes with the appropriate number of corresponding days per episode. With 1000 or fewer strategy updates, the solution quality notably decreases (Table~\ref{table:batch_size}). We use 1000 episodes of length 1000 -- yielding both adherence and fairness over 0.90 -- in the remaining experiments.

\begin{table}
	\scriptsize
	\centering
	\caption{Comparison for various numbers of batches and episode lengths $m$. A balance between both yields the best results.}
	\label{table:batch_size}
	\begin{tabular}{ l l l l l}
		\toprule
		\# batches & $m$ & Utilization & Nash adherence & Fairness\\	
		\midrule	
		100000 & 10 &1.00 (1.00) & 0.95 (0.96) & 0.88 (0.91) \\
		10000 & 100 &0.99 (0.99) & 0.93 (0.93) & 0.90 (0.85) \\
		1000 & 1000 &0.99 (0.99) & 0.92 (0.94) & 0.93 (0.92) \\
		100 & 10000 &0.89 (0.99) & 0.78 (0.87) & 0.82 (0.92) \\
		10 & 100000 &0.71 (0.90) & 0.51 (0.66) & 0.65 (0.83) \\
		\bottomrule  
	\end{tabular}
\end{table}

\subsubsection{Learning rates}
We test a variety of learning rates $\alpha \in \{0.1, 0.01, 0.001, 0.0001\}$. Higher learning rates are more responsive to changes in the system, but also display worse convergence behavior
(Table~\ref{table:learning rates}). Learning rates 0.1 and 0.01 are unstable due to exploding gradients. We find that 0.001 generally provides the best average results, but lower learning rates yield better results after 100,000 iterations. As we perform online learning that requires responsiveness as well as quality, we keep $\alpha=0.001$. 

\begin{table}
	\scriptsize
	\centering
	\caption{Comparison for various learning rates. Low learning rates display better convergence, but worse results on average.}
	\label{table:learning rates}
	\begin{tabular}{ l l l l }
		\toprule
		$\alpha$ & Utilization & Nash adherence & Fairness\\	
		\midrule	
		0.1      & N/A & N/A & N/A \\
		0.01     & N/A & N/A & N/A \\
		0.001    & 0.99 (0.99) & 0.92 (0.94) & 0.92 (0.92) \\
		0.0001   & 0.99 (1.00) & 0.94 (0.97) & 0.73 (0.96) \\
		0.00001  & 0.94 (1.00) & 0.78  (0.98) & 0.78 (0.93) \\
		\bottomrule  
	\end{tabular}
\end{table}

\subsubsection{Initial standard deviations}
We test initial standard deviations $\sigma_{\boldsymbol{\theta}}^0$ ranging from 0.01 to 2 (results in Table~\ref{table:sigma}). Overall, results are quite close. Lower initial standard deviations encourage less exploration and increases time to converge, whereas a larger deviation may cause stronger fluctuations that impact fairness in particular. A standard deviation of 0.1 yields the best results for this particular setting. We note that -- regardless of initialization -- standard deviations converge to similar, relatively low values eventually. However, there seemingly is a benefit to preserve some randomness in the actions, as standard deviations do not completely converge to 0.

\begin{table}
	\scriptsize
	\centering
	\caption{Comparison for $\sigma_{\boldsymbol{\theta}}^0$ initialization. Results are fairly close, but $\sigma_{\boldsymbol{\theta}}^0=0.1$ strikes a good balance between fairness and Nash adherence.}
	\label{table:sigma}       
	\begin{tabular}{ l l l l  }
		\toprule
		$\sigma_{\boldsymbol{\theta}}^0$ & Utilization &  Nash adherence & Fairness \\
		\midrule	
		0.01& 0.99 (0.99) & 0.89 (0.94) & 0.86 (0.88) \\
		0.1 & 0.99 (0.99) & 0.91 (0.93) & 0.92 (0.93) \\
		0.5 & 0.99 (0.99) & 0.91 (0.86) & 0.86 (0.82) \\
		1.0 & 0.99 (1.00) & 0.91 (0.91) & 0.95 (0.95) \\
		1.5 & 0.99 (0.99) & 0.90 (0.92) & 0.89 (0.86) \\
		2.0 & 0.99 (0.99) & 0.91 (0.91) & 0.85 (0.94) \\
		\bottomrule  
	\end{tabular}
\end{table}

\subsubsection{Actor networks architectures}\label{sssec:actor_networks_architectures} 
We now address actor network architectures. We test a setting without hidden layer (i.e., a linear model) and 1 to 3 layers with 5 to 30 nodes each. Adding nodes and layers may capture more complicated functions, but take both additional time and observations to train. Table~\ref{table:network_architectures} shows the key results; the full table of experiments is displayed in Appendix~B. First, we find that performance is fairly stable across architectures. Some sub-par convergence behavior is noted for both the smallest (5 nodes) and largest (25 or 30 nodes) networks, although all converge to good (0.90+) metric scores eventually. Second, we find that computational times are fairly similar for various network sizes, with ~10\% difference between network configurations. Third, the linear approximation schemes outperforms the neural network architectures with respect to eventual performance, but does notably worse on average fairness. We continue with an actor network consisting of one layer with 20 nodes. As Rolnick \& Tegmark \cite{rolnick2017} point out, relatively small neural networks suffice for many real-world problems. For our system, this notion appears to apply as well.

\begin{table}
	\scriptsize
	\centering
	\caption{Comparison of various actor network architectures. The 1-layer network with 20 nodes yields the best performance on average. The linear architecture performs best on Nash adherence but worst on fairness.}
	\label{table:network_architectures}
	\begin{tabular}{ l l l l  }
		\toprule
		Architecture & Utilization &  Nash adherence & Fairness \\
		\midrule	
		Linear & 0.99 (1.00) & 0.96 (0.97) & 0.86 (0.97)  \\
		1L 20N & 0.99 (1.00) & 0.91 (0.91) & 0.95 (0.95) \\
		2L 20N & 0.99 (0.99) & 0.89 (0.91) & 0.92 (0.94) \\
		3L 20N & 0.99 (0.99) & 0.89 (0.93) & 0.92 (0.96) \\
		\bottomrule  
	\end{tabular}
\end{table}

\subsubsection{Policy gradient algorithms}
We test five variants of the policy gradient algorithm: (i) vanilla policy gradient, (ii) policy gradient with baseline, (iii) actor-critic with Q-value, (iv) TD(1) and (v) actor-critic advantage function. For the critic networks we use the same architecture as for the actor networks; one layer with 20 nodes. The key results are summarized in Table~\ref{table:policy_gradient_algorithms}. The policy gradient algorithms consistently outperform the actor-critic methods; especially TD(1) and Advantage Value suffer from premature convergence and perform very poorly. As target values keep changing over time, simultaneously updating the strategies and value functions is notoriously hard. An incorrect Q-value leads to poor strategies and vice versa, as reflected in the performance. This observation is in line with Grondman \textit{et al.} \cite{grondman2012}, stating that actor-critic methods are less suitable for highly non-stationary environments. We therefore prefer policy gradient algorithms; adding a baseline does not have a major impact on performance.

\begin{table}
	\scriptsize
	\centering
	\caption{Comparison of policy gradient algorithms. The actor-critic algorithms suffer from poor convergence.}
	\label{table:policy_gradient_algorithms}       
	\begin{tabular}{ l l l l  }
			\toprule
			Algorithm & Utilization &  Nash adherence & Fairness \\
			\midrule	
			Policy gradient             & 0.99 (0.99) & 0.92 (0.94)   & 0.93 (0.92) \\
			Policy gradient w. baseline & 0.99 (0.99) & 0.90 (0.91)   & 0.95 (0.96) \\
			Q-value                     & 0.82 (0.98) & 0.69 (0.86)   & 0.66 (0.81) \\
			TD(1)                       & 0.09 (0.00) & 0.07 (0.00)   & 0.08 (0.00) \\
			Advantage value             & 0.98 (1.00) & 0.00 (0.00)   & 0.00 (0.00) \\
			\bottomrule  
	\end{tabular}
\end{table}

\subsubsection{Critic network}
Finally, we test various architectures for the critic network, using either a linear approximation scheme or 1 to 3 hidden layers. The one-layer network outperforms all alternatives (Table~\ref{table:Q-value architectures}). Furthermore, it is confirmed again that the actor-critic approaches do not achieve consistent results.

\begin{table}
	\scriptsize
	\centering
	\caption{Comparison of critic network architectures. Similar to the actor networks, a 1-layer network with 20 nodes yield the best results. Overall, performance when adding a critic network is subpar.}
	\label{table:Q-value architectures}      
	\begin{tabular}{ l l l l  }
		\toprule
		Architecture  & Utilization &  Nash adherence & Fairness \\
		\midrule	
		Q-value linear& 0.95 (0.95) & 0.05 (0.17) & 0.00 (0.00) \\
		Q-value 1L20N & 0.82 (0.98) & 0.69 (0.86) & 0.66 (0.81) \\
		Q-value 2L20N & 0.06 (0.57) & 0.05 (0.51) & 0.00 (0.00) \\
		Q-value 3L20N & 0.17 (0.91) & 0.13 (0.77) & 0.57 (0.30) \\
		\bottomrule  
	\end{tabular}
\end{table}

\subsubsection{Summary}
We summarize the main conclusions of our exploration. Due to the highly non-stationary environment, convergence to stable solutions is relatively slow. Policy gradient algorithms without a critic deal much better with this non-stationarity, responding more quickly to environmental changes. With the exception of critic-based solutions, we find solutions to be fairly stable across test settings. Still, there is an inherent trade-off between responsiveness and solution quality. Deep neural networks and low learning rates yield good results for static targets (reflected in the end-of-horizon results), but average performance worsens in dynamic settings. For our online setting, we therefore prefer a shallow neural network and a moderate learning rate. The best setting found consists of an equal balance between batch size and number of episodes (both 1000), a learning rate of 0.001, an initial standard deviation of 0.1, and an actor-network consisting of one layer with 20 neurons.

\subsection{Analysis Case I}\label{ssec:analysis_case1} 
In Section~\ref{ssec:preliminary} we discussed the elementary system behavior and determined appropriate parameters. We now further delves into the competitive elements of the environment, with both agents aiming to obtain the best deal for themselves by making strategic decisions. In addition to the system metrics, we therefore also consider agent-specific metrics, in particular their average rewards. For this purpose, we present the reward shares for carrier and shipper instead of fairness; note that the remaining share goes to the broker. 

We explore whether agents can outperform their counterpart with a deviating strategy. For convenience we mostly keep the behavior of the carrier fixed and alter that of the shipper; similar results could be obtained the other way around. This section addresses the following aspects: (i) asymmetric learning rates, (ii) penalty function, (iii) actor network architecture, (iv) actor-critic algorithm, (v) bias initialization and (vi) standard deviation initialization. Based on the obtained insights, we then run a number of experiments in which both shipper's and carrier's strategies vary based on their risk profile, trying to determine the best response to the opponent's strategy.

\subsubsection{Asymmetric learning rates}
The learning rate determines how responsive the strategy updates are to new information. If an agent uses a higher learning rate than it opponents, it could adapt faster to a new environment. We find that, to a certain degree, higher learning rates yields better results (Table~\ref{table:asymmetric learning rates}), with the shipper outearning the carrier by up to 9\%. For learning rates of 0.01 and higher, we did not consistently find stable solutions.


\begin{table}
	\scriptsize
	\centering
	\caption{Comparison of asymmetric learning rates. The carrier has a fixed learning rate of 0.001. Higher learning rates yield an advantage in terms of reward share.}
	\label{table:asymmetric learning rates}      
	\begin{tabular}{ l l l l l }
		\toprule
		Learning rate ($S$) & Utilization &  Nash adherence & Reward share ($S$) & Reward share ($C$) \\
		\midrule	
		0.0001 & 0.99 (1.00) & 0.92 (0.95) & 0.33 (0.40) & 0.51 (0.56) \\
		0.0005 & 0.99 (0.99) & 0.87 (0.91) & 0.37 (0.45) & 0.47 (0.50) \\
		0.001  & 0.99 (0.99) & 0.92 (0.93) & 0.40 (0.43) & 0.43 (0.48) \\
		0.002  & 0.99 (0.99) & 0.91 (0.92) & 0.43 (0.44) & 0.40 (0.48)  \\
		0.005  & 0.99 (0.99) & 0.91 (0.94) & 0.43 (0.48) & 0.39 (0.47) \\
		0.01   & N/A         & N/A         & N/A         & N/A   \\
		\bottomrule  
	\end{tabular}
\end{table}

\subsubsection{Penalty function}
The slope of the penalty functions (Equation~\eqref{eq:rewardshipperpenalty}-\eqref{eq:rewardcarrierpenalty}) govern the magnitude of the artificial penalty incurred when jobs are not shipped. By default both have a slope of 1, such that missed rewards are weighted the same as realized rewards. Here we test a number of alternative slopes, ranging from 0 (i.e., no penalty) to 5 (high risk aversion). We find that lower penalties average yield strategies with higher realized rewards (Table~\ref{table:penalty functions}). A shipper without a penalty function obtains up to 68\% of the revenues, versus only 26\% for the carrier (the remainder going to the broker). Nash adherence and utilization are only marginally affected, but do go down slightly with risk-seeking behavior.




\begin{table}
	\scriptsize
	\centering
	\caption{Comparison of asymmetric penalty functions. The carrier has a fixed penalty of 1. Lower penalties considerably benefit the shipper.}
	\label{table:penalty functions}      
	\begin{tabular}{ l l l l l }
		\toprule
		Pen. slope ($S$) & Utilization &  Nash adherence & Reward share ($S$) &  Reward share ($C$) \\
		\midrule	
		0   & 0.98 (0.98) & 0.92 (0.93) & 0.64 (0.68) & 0.21 (0.26) \\
		0.5 & 0.99 (0.99) & 0.93 (0.94) & 0.49 (0.51) & 0.34 (0.42) \\
		1.0 & 0.99 (0.99) & 0.92 (0.93) & 0.40 (0.43) & 0.43 (0.48) \\
		2.0 & 0.99 (0.99) & 0.93 (0.94) & 0.32 (0.38) & 0.51 (0.56) \\
		5.0 & 0.99 (1.00) & 0.93 (0.94) & 0.26 (0.33) & 0.58 (0.61) \\
		\bottomrule  
	\end{tabular}
\end{table}

\subsubsection{Linear approximation vs actor network}
By default we use actor networks with 1 layer and 20 neurons. In this experiment, we test whether adopting a linear approximation scheme by the shipper yields a benefit or disadvantage, being more responsive but less powerful. We find that, in this setting, the average reward shares are 32\% (was 40\%) for the shipper and 53\% (was 43\%) for the carrier. This substantial difference implies that the neural network aids agents in making better decisions.
	
\subsubsection{Q-value vs. policy gradient}
Section~\ref{ssec:preliminary} already indicated that algorithms relying on value function approximation performs much worse than those relying on reward observations. We now test a setting where only the shipper uses an actor-critic method, checking once more whether the additional expressive power of the critic network could give an advantage. Again, the results are unconvincing. Nash adherence is only 65\%, and on average the shipper earns a negative reward. Although some runs yield reasonable outcomes for the shipper, others yield unstable solutions in which shipping agreements are rare.

\subsubsection{Initial bias}
In Section~\ref{ssec:preliminary} we assumed that an agent's initial bid (ask) equals its willingness to pay (marginal transport costs). However, this directly reveals their highest bid; in reality they may start with a lower bid (higher ask) and update until finding a balance. Thus, we conduct an experiment in which the shipper starts with lower bids; the initial average bid is set at 0, 0.5, 1.0, 1.5, or 2.0. Naturally, the bias weight is updated over time as before. The results (Table~\ref{table:bias_initialization}) show that starting out with a low bid is clearly beneficial, with the bid-ask pairs converging to equilibria that are considerably better for the shipper.

\begin{table}
	\scriptsize
	\centering
	\caption{Comparison of varying initial bias. The carrier's initial bias is set at $\mu_{\boldsymbol{\theta}}^{C,0}=1$. Low bias initializations strongly improve performance for the shipper.}
	\label{table:bias_initialization}      
	\begin{tabular}{ l l l l l }
		\toprule
		Bias $\mu_{\boldsymbol{\theta}}^{S,0}$  & Utilization &  Nash adherence & Reward share shipper &  Reward share carrier \\
		\midrule	
		0.0 & 0.99 (0.99) & 0.93 (0.92) & 0.81 (0.62) & 0.23 (0.32) \\
	    0.5 & 0.99 (0.99) & 0.93 (0.93) & 0.75 (0.60) & 0.25 (0.34) \\
		1.0 & 0.99 (0.99) & 0.94 (0.95) & 0.66 (0.62) & 0.29 (0.33) \\
		1.5 & 0.99 (0.99) & 0.93 (0.93) & 0.51 (0.49) & 0.39 (0.45) \\
		2.0 & 0.99 (0.99) & 0.92 (0.93) & 0.40 (0.43) & 0.43 (0.48) \\
		\bottomrule  
	\end{tabular}
\end{table}

\subsubsection{Initial standard deviation}
A higher initial standard deviation than the opponent enables exploring more, yet also increases the risk of ending up in unfeasible solution regions. We test for $\sigma_{\boldsymbol{\theta}}^{S,0} \in \{0.01,0.1,0.5,1,1.5,2.0\}$. This only entails the initial standard deviations; we find them all converging to similar values regardless of initialization. The experiments show a clear disadvantage when the shipper uses lower standard deviations (Table~\ref{table:sigma_initialization}). For higher standard deviations performance also decreases, although not as sharply. There appears no imminent benefit to alter the standard deviation, as the default value yields the best outcome.

\begin{table}
	\scriptsize
	\centering
	\caption{Comparison of varying initial st. dev. The carrier's initial st. dev. is set at $\sigma_{\boldsymbol{\theta}}^{C,0}=1$. Middle-range exploration yields better results for the shipper.}
	\label{table:sigma_initialization}      
	\begin{tabular}{ l l l l l }
		\toprule
		St. dev. $\sigma_{\boldsymbol{\theta}}^{S,0}$ & Utilization &  Nash adherence & Reward share shipper &  Reward share carrier \\
		\midrule	
		0.01 & 0.99 (0.99) & 0.93 (0.93) & 0.32 (0.44) & 0.52 (0.50) \\
		0.1  & 0.99 (0.99) & 0.93 (0.94) & 0.36 (0.46) & 0.49 (0.49) \\
		0.5  & 0.99 (0.99) & 0.92 (0.93) & 0.40 (0.47) & 0.44 (0.47) \\
		1.0  & 0.99 (0.99) & 0.92 (0.93) & 0.40 (0.43) & 0.43 (0.48) \\
		1.5  & 0.99 (0.99) & 0.92 (0.93) & 0.38 (0.42) & 0.45 (0.51) \\
		2.0  & 0.99 (0.99) & 0.92 (0.92) & 0.35 (0.41) & 0.49 (0.52) \\
		\bottomrule  
	\end{tabular}
\end{table}

\subsubsection{Shipper vs. carrier}\label{sssec:risk_profiles_case1}
After completing the experiments varying only the shipper's behavior, we apply the acquired insights in a number of experiments in which both agents adopt strategy settings according to a certain risk appetite. We define three profiles -- risk-seeking, risk-neutral, risk-averse -- varying in initial bias, penalty slope and learning rate (see Table~\ref{table:risk profiles}).

\begin{table}
	\scriptsize
	\centering
	\caption{Risk profiles for carrier and shipper. A risk-averse profile is aimed at finding feasible solutions, a risk-seeking profile at maximizing individual gains.}
	\label{table:risk profiles}      
	\begin{tabular}{ l l l }
		\toprule
		Risk profile & Carrier                & Shipper  \\
		\midrule	
		Risk-seeking & Init. Bias = 1         & Init. Bias = 2 \\
		             & Penalty slope = 2      & Penalty slope = 2 \\
		             & Learning rate = 0.005  & Learning rate = 0.005 \\
		Risk-neutral & Init. Bias = 1         & Init. Bias = 2 \\
					 & Penalty slope = 1      & Penalty slope = 1 \\
				 	 & Learning rate = 0.001  & Learning rate = 0.001 \\
		Risk-averse  & Init. Bias = 1         & Init. Bias = 2 \\
					 & Penalty slope = 2      & Penalty slope = 2 \\
					 & Learning rate = 0.0001 & Learning rate = 0.0001 \\
		\bottomrule  
	\end{tabular}
\end{table}

\begin{figure}[ht] 
	\begin{minipage}{.5\textwidth}
		\includegraphics[width=\textwidth]{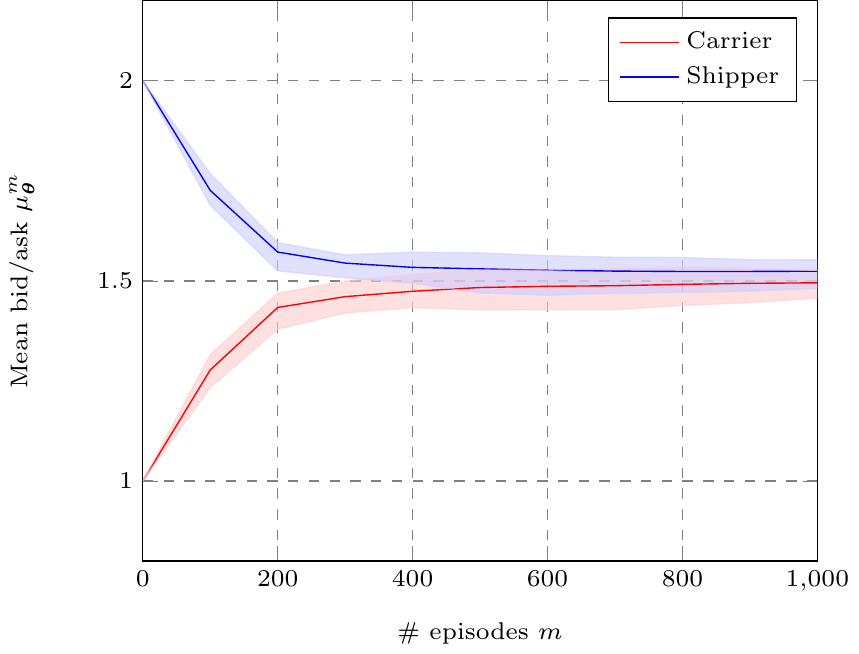}
		\caption{Carrier RA and shipper RA}
		\label{fig:2a}       
	\end{minipage}
	\begin{minipage}{.5\textwidth}
		\includegraphics[width=\textwidth]{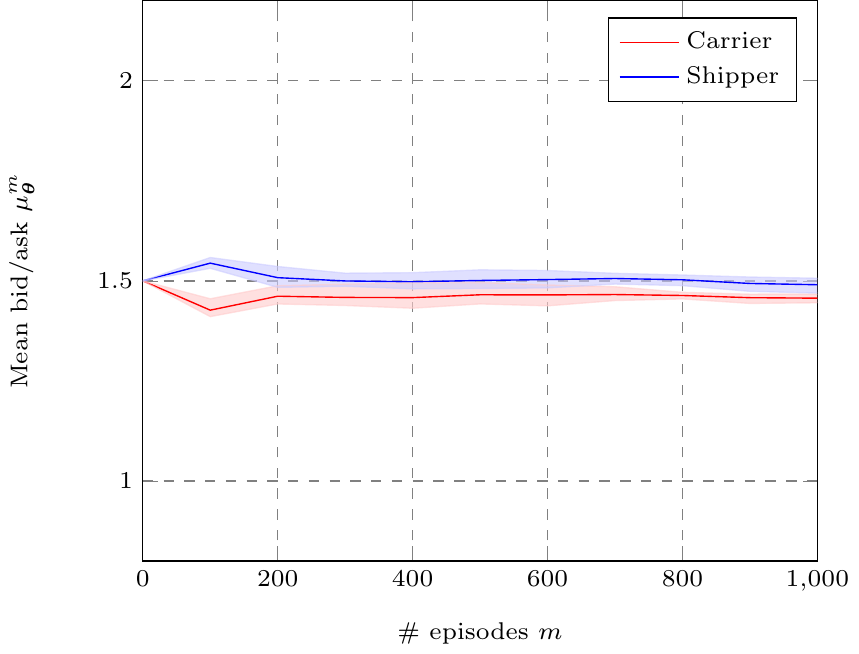}
		\caption{Carrier RN and shipper RN}
		\label{fig:2b}       
	\end{minipage}
	\begin{minipage}{.5\textwidth}
		\includegraphics[width=\textwidth]{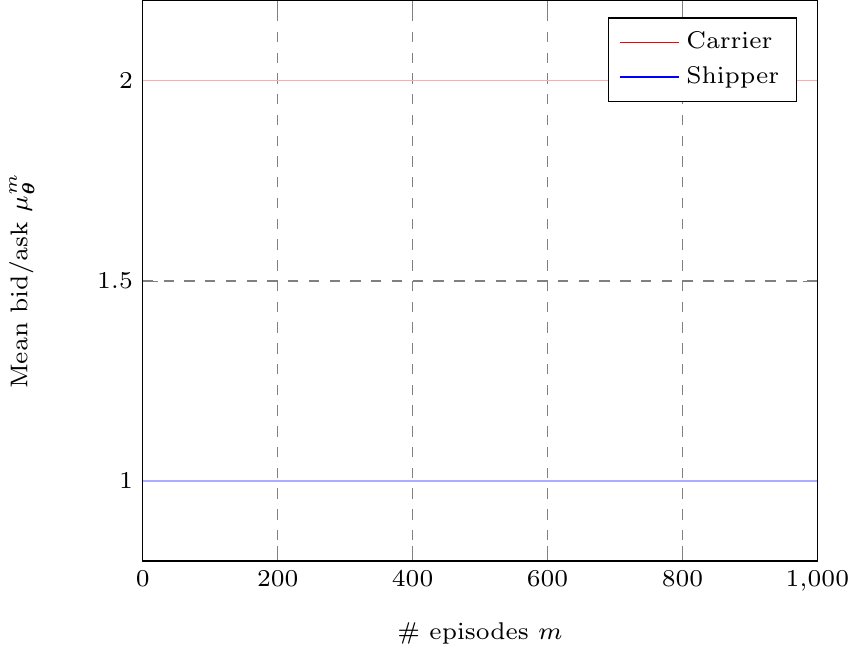}
		\caption{Carrier RS and shipper RS}
		\label{fig:2c}       
	\end{minipage}
	\begin{minipage}{.5\textwidth}
		\includegraphics[width=\textwidth]{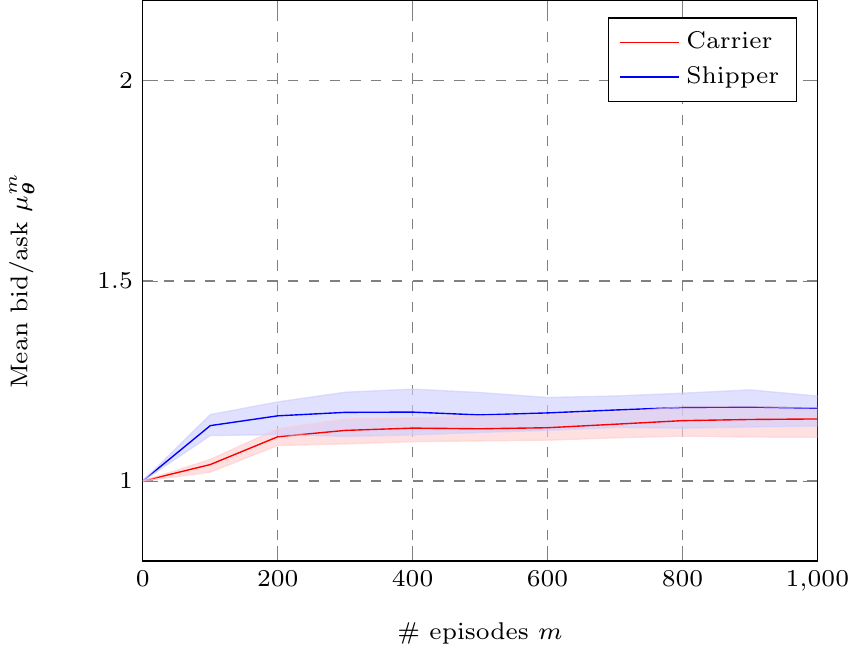}%
		\caption{Carrier RA and shipper RS}
		\label{fig:2d}       
	\end{minipage}
\end{figure}

Using the risk profiles, we perform nine experiments pitting each pair of these profiles against one another. The results are shown in Table~\ref{table:risk profile experiments}. Overall, the results clearly show that the more risk-seeking agent consistently obtains better rewards. Figures~\ref{fig:2a}-\ref{fig:2d} illustrate the dynamics between agents with different risk profiles for four settings. The bid-ask prices converge to equilibria that are more favorable to the agent with the higher risk tolerance. However, we note that, when both agents use a risk-seeking profile, they stick to their initial bias and never explore feasible solution regions, partially due to the absence of penalties. This shows that some form of reward shaping is desirable for a functioning system. We tested some variants with two risk-seeking agents (e.g., with small penalty slopes or larger standard deviations), but they tend to be either unstable or not converging.

In terms of system performance, we find no major differences, with utilizations close to 1 and Nash adherence generally between 0.90 and 0.95. Occasionally systems with a risk-seeking agent fail to find equilibria though, (temporarily) resulting in non-functioning transport markets. From a game-theoretical perspective, Appendix~D shows that when the opponent is risk-seeking, the best response is to be either risk-averse (for carrier) or risk-neutral (for shipper).
 
\begin{table}
	\scriptsize
	\centering
	\caption{Performance for various risk profiles. Each cell report 'reward share carrier/reward share shipper [utilization]'. Risk-seeking behavior if generally rewarded, but not if both agents do the same.}
	\label{table:risk profile experiments}   
	\begin{tabular}{ l| l l l  }
		\hline
		\backslashbox{Carrier}{Shipper} & RS &  RN & RA \\
		\hline	
		RS & N/A [0.00]       & 0.81/0.18 [0.85] & 0.82/0.14  [0.95]\\
		RN & 0.25/0.75 [0.97] & 0.46/0.49 [0.99] & 0.62/0.28 [0.99] \\
		RA & 0.11/0.85 [0.98] & 0.26/0.63 [0.99] & 0.41/0.40 [1.00]  \\
		\hline  
	\end{tabular}
\end{table}

\subsubsection{Summary Case I experiments}
From the experiments, we can draw a number of conclusions. We reconfirm that value function approximation lead to poor performance and that actor networks outperform linear approximations. Furthermore, we find that risk-seeking behavior generally pays off. By responding faster to market fluctuations (higher learning rate), placing lower penalty weights on failed shipments, and setting a bold initial opening bid/ask, an agent may obtain an edge over the competitor. However, we also find evidence of destabilizing effects and poor market efficiency due to risk-seeking behavior. Without central regulation or communication between agents, risk-seeking behavior might have grave consequences; game theory may help to intelligently respond to the other agent's strategy. As the performance results can be quite substantial, these findings imply that algorithmic optimization is a crucial step to successfully participate in a decentralized freight transport market.









\subsection{Analysis Case II}\label{ssec:analysis_case2} 
This section analyzes the second experimental setting, which generates up to 10 jobs per day with varying properties in terms of volume, due date and distance. Case II serves as a proof-of-concept and to measure performances in a more realistic setting. To recap: we now consider a setting with 125 distinct job types (varying in due date, volume and distance), with order accumulation possible up to 60 jobs. With a maximum daily volume of 50 and a transport capacity of either 40 (somewhat scarce) or 300 (abundant), we evaluate instances that favor either the carrier or the shipper. Following Section~\ref{sssec:risk_profiles_case1}, we again utilize the three risk profiles, starting by using the same profiles for both agents. The bias initialization procedure is essentially similar as before, but notationally more involved and therefore formalized in Appendix~C.

The initial results of running the three paired risk profiles illustrate the increased challenges of the stochastic case (Table~\ref{table:analysis case 2}). The risk-seeking and risk-neutral profiles do not consistently yield stable solutions. For the risk-averse profiles, utilization ranges from 67\% to 89\%, indicating an increased difficulty to find suitable transport matches. Furthermore, adherence to the Nash equilibrium drops to 65\%, revealing substantial market inefficiencies. Bid-ask spreads are higher due to the large uncertainty in the market; the broker now takes 35\% of the market value. Finally, fairness and the reward shares indicate unbalances in the market. However, recall from Equations~\eqref{eq:rewardshipperpenalty}-\eqref{eq:rewardcarrierpenalty} that penalty functions are not symmetric.
\begin{table}[h!]
	\scriptsize
	\centering
	\caption{Results Case II per risk profile, with both agents using the same profile. Compared to Case I, Nash adherence drops substantially. Risk-seeking profiles do not achieve stable results.}
	\label{table:analysis case 2}      
	\begin{tabular}{ l l l l l l}
		\toprule
		Risk profile & Utilization &  Nash adherence & Fairness&Reward share  &  Reward share  \\
		(capacity) &  &   & &shipper & carrier \\
		\midrule	
		Risk-seeking ($\zeta^C=40$) & N/A & N/A & N/A & N/A & N/A\\
		Risk-seeking ($\zeta^C=300$)   & N/A &N/A & N/A &N/A & N/A\\
		Risk-neutral ($\zeta^C=40$)  & 0.86 (0.86) & 0.53 (0.43) & 0.14 (0.16) & 0.34 (0.01) & 0.23 (0.20)\\
		Risk-neutral ($\zeta^C=300$)  &  N/A &  N/A & N/A &  N/A & N/A  \\
		Risk-averse ($\zeta^C=40$)  & 0.89 (0.80) & 0.65 (0.67) & 0.13 (0.14) & 0.05 (0.13) & 0.74 (0.73)\\
		Risk-averse ($\zeta^C=300$)  & 0.67 (0.76) & 0.65 (0.22) & 0.23 (0.04) & 0.06 (0.12) & 0.73 (0.73)\\
		\bottomrule  
	\end{tabular}
\end{table}

To improve system performance, we execute a number of follow-up experiments. Using the insights obtained from the Case I experiments, we vary the initial bias and sigma, the learning rate, and the penalty slope). As only the risk-averse profiles yield consistent results, we consider several variants of that profile. Table~\ref{table:risk-averse variants} shows that most modifications are unsuccessful in achieving notable improvements, yet using a risk-neutral bias initialization (i.e., halfway expected willingness to pay and expected transport costs) strongly improves performance. We find a utilization of 0.98-0.99 and a Nash adherence of 0.84-0.87. Although fairness remains low (recall that fairness is calculated at the individual job level), for overall profits the shares of carrier and shipper differences are less pronounced (0.54/0.39 and 0.60/0.32).

\begin{table}
	\scriptsize
	\centering
	\caption{Results Case II for variants of risk-averse profiles. Setting an appropriate initial bias has by far the largest positive impact on system performance.}
	\label{table:risk-averse variants}      
	\begin{tabular}{ l l l l l l}
		\toprule
		Variant & Utilization &  Nash adherence & Fairness&Reward share  &  Reward share  \\
		(capacity) &  &   & &shipper & carrier \\
		\midrule	
		lr=0.00001 ($\zeta^C=40$)  & 0.98(0.98) & 0.41 (0.61) & 0.09 (0.13) & 0.06 (0.03) & 0.43 (0.76)\\
		lr=0.00001 ($\zeta^C=300$)   & 0.98(0.98)  & 0.42 (0.59) & 0.09 (0.13)  & 0.05 (0.00) & 0.46 (0.77) \\
		$\sigma=10$($\zeta^C=40$)  & 0.99 (1.00) & 0.53 (0.54) & 0.13 (0.12) & 0.30 (0.38) & 0.36 (0.32) \\
		$\sigma=10$ ($\zeta^C=300$)  & 0.99 (0.99)& 0.53 (0.60) & 0.12 (0.11) &  0.31 (0.42) & 0.37 (0.34)\\
		Pen. slope = 0 ($\zeta^C=40$)  & 0.01 (0.00) & 0.01 (0.00) & N/A & N/A  &N/A\\
		Pen. slope = 0 ($\zeta^C=300$) & 0.13 (0.10) & 0.12 (0.13)  & 0.22 (0.23) & 0.87 (N/A)  & -0.02 (N/A) \\
		Pen. slope = 1 ($\zeta^C=40$)  & 0.97 (0.95) & 0.62 (0.69) & 0.13 (0.14) & -0.03 (-0.08) & 0.81 (0.93)\\
		Pen. slope = 1 ($\zeta^C=300$)  & 0.99 (1.00) &  0.60 (0.64)  & 0.13 (0.14) & -0.26 (-0.76)  & 0.78 (0.89)  \\
		RN bias ($\zeta^C=40$)   & 0.99 (0.99) & 0.87 (0.91) & 0.14 (0.16) & 0.54 (0.56) & 0.39 (0.40)\\
		RN bias ($\zeta^C=300$)  & 0.98 (0.98) & 0.84 (0.89) & 0.14 (0.15) & 0.60 (0.63) & 0.32 (0.31)\\
		RS bias ($\zeta^C=40$)   & N/A & N/A  & N/A & N/A &N/A\\
		RS bias ($\zeta^C=300$)  & N/A & N/A  & N/A & N/A &N/A\\
		\bottomrule  
	\end{tabular}
\end{table}

The system effectiveness of starting with a fair bid/ask an important result. However, it is not necessarily individual rational. We see that the shipper's performance improves drastically (the reward share increases from 0.05 to 0.54), but for the carrier the reward share drops from 0.74 to 0.39. The increased number of jobs shipped does not compensate for that. As a final experiment, we therefore again test the impact of asymmetric agent profiles, determining the individually rational bias initializations. Although we describe biases in terms of risk, the learning rate and penalty function remain risk-averse. The results are shown in Table~\ref{table:risk profile experiments case2 (cap40)} ($\zeta^C=40$) and Table~\ref{table:risk profile experiments case2 (cap300)} ($\zeta^C=300$). As before, risk-seeking approaches generally improve individual performance, although the results are less congruous and clear-cut as for the deterministic case.

\begin{figure}[ht] 
	\begin{minipage}{.5\textwidth}
		\includegraphics[width=\textwidth]{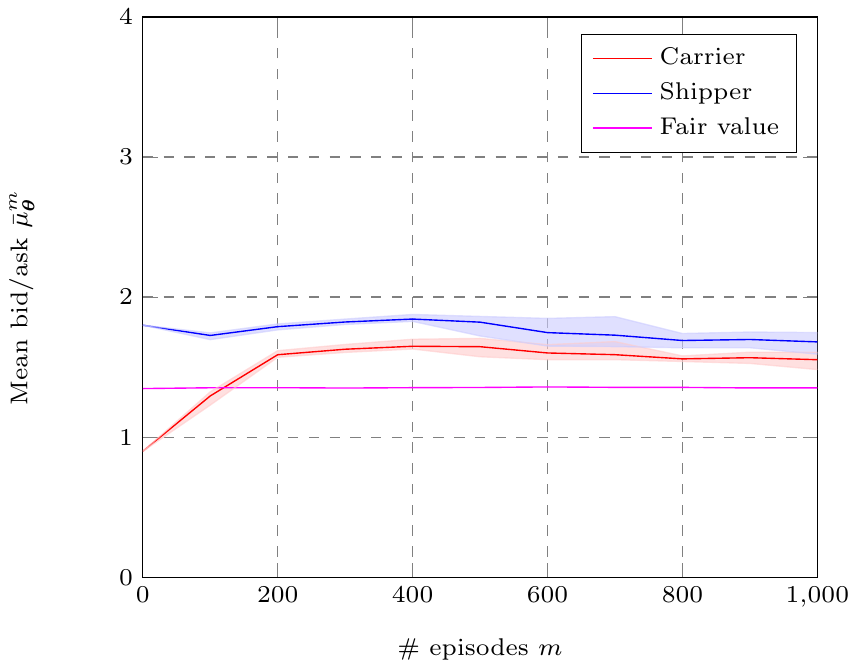}
		\caption{Carrier RA bias and shipper RA bias ($\zeta^C=40$)}
		\label{fig:3a}       
	\end{minipage}
	\begin{minipage}{.5\textwidth}
		\includegraphics[width=\textwidth]{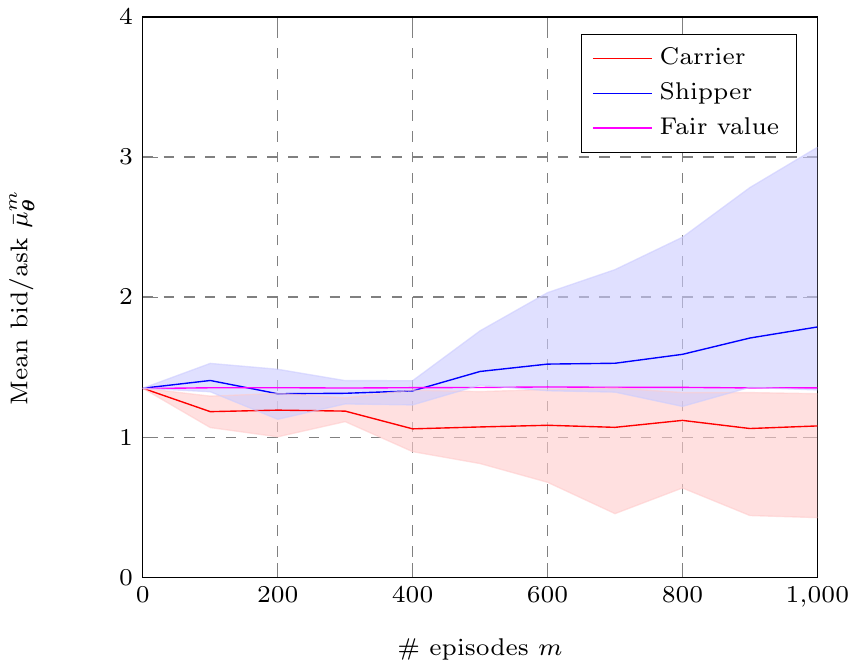}
		\caption{Carrier RN bias and shipper RN bias ($\zeta^C=40$)}
		\label{fig:3b}       
	\end{minipage}
	\begin{minipage}{.5\textwidth}
		\includegraphics[width=\textwidth]{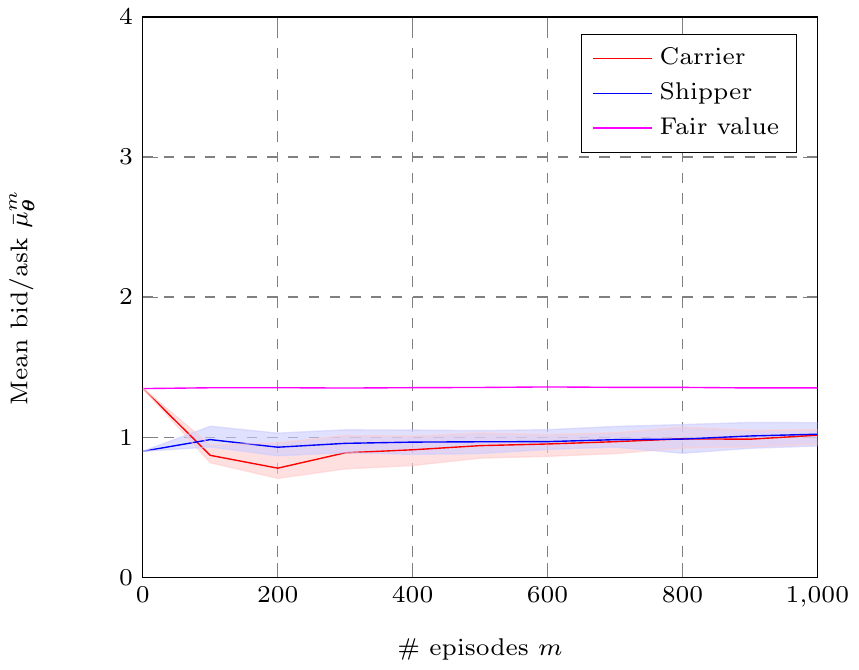}
		\caption{Carrier RN bias and shipper RS bias ($\zeta^C=40$)}
		\label{fig:3c}       
	\end{minipage}
	\begin{minipage}{.5\textwidth}
		\includegraphics[width=\textwidth]{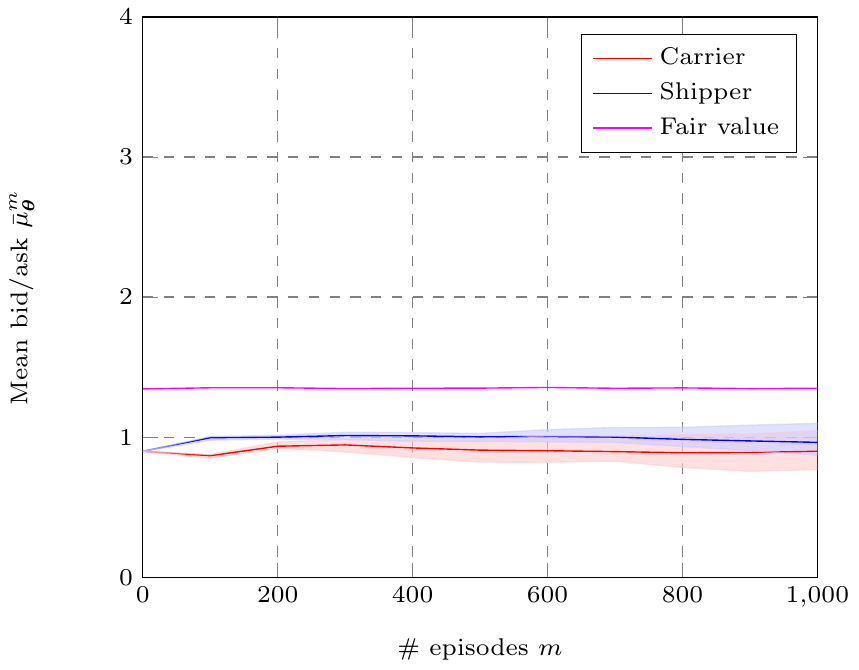}%
		\caption{Carrier RA bias and shipper RS bias ($\zeta^C=300$)}
		\label{fig:3d}       
	\end{minipage}
\end{figure}

We discuss the best response of individual agents, considering the bias initialization as a game on itself. Formally, we achieve Nash equilibria\footnote{The Nash equilibrium discussed here (for the game of initializing the bias) should not be convoluted with the Nash equilibrium for the bargaining game as mentioned earlier.} -- in a non-cooperative context \cite{nash1951} -- when the shipper adopts the risk-seeking bias initialization and the carrier uses the risk-neutral bias ($\zeta^C=40$) or risk-averse bias ($\zeta^C=300$), see Appendix~D for details and payoff matrices. With a somewhat less rigorous interpretation, the payoffs imply that if one agent adopts a bold bias, the opponent should settle for lower individual gains. If both agents use a bold bid- or ask strategy, systems do not converge properly and individual gains are minimal as a result. Figures~\ref{fig:3a}-\ref{fig:3d} highlight results for four risk profile pairs, including the two Nash equilibria.

\begin{table}
	\scriptsize
	\centering
	\caption{Performance for various bias initializations, $\zeta=40$. Each cell report 'reward share carrier/reward share shipper [utilization]'. Note that the agents always use the risk-averse learning rate and penalty function; only the bias initializations deviate. }
	\label{table:risk profile experiments case2 (cap40)}     
	\begin{tabular}{ l| l l l  }
		\hline
		\backslashbox{Carrier}{Shipper} & RS Bias &  RN Bias & RA Bias \\
		\hline	
		RS Bias & N/A [0.0]        & 0.18/0.69 [0.82] & 0.72/0.08 [0.99] \\
		RN Bias & 0.02/0.91 [0.63] & 0.39/0.54 [0.99] & 0.88/-0.01 [0.98] \\
		RA Bias & 0.00/0.94 [0.96] & 0.35/0.51 [0.92] & 0.74/0.05 [0.89]  \\
		\hline  
	\end{tabular}
\end{table}

\begin{table}
	\scriptsize
	\centering
	\caption{Performance for various bias initializations, $\zeta=300$. Each cell report 'reward share carrier/reward share shipper [utilization]'. Note that the agents always use the risk-averse learning rate and penalty; only the bias initializations deviate.}
	\label{table:risk profile experiments case2 (cap300)}  
	\begin{tabular}{ l| l l l  }
		\hline
		\backslashbox{Carrier}{Shipper} & RS &  RN & RA \\
		\hline	
		RS & -0.02/0.87 [0.13]  & 0.09/0.73 [0.92] & 0.69/0.21 [0.99]  \\
		RN & -0.09/0.98 [0.78]  & 0.32/0.60 [0.98] & 0.95/-0.08 [0.99]  \\
		RA & 0.01/0.89 [0.98]   & 0.32/0.53 [0.92] & 0.73/0.06 [0.67]  \\
		\hline  
	\end{tabular}
\end{table}

To conclude, in the stochastic case (with 125 job types and uncertain numbers of jobs) stability is harder to achieve than in the deterministic case, requiring more cautious updating strategies. The main determinant for the agents' individual gains is their initial bias (colloquially the opening bid/ask), which to a large degree determines the eventual convergence. Fairness is  hard to achieve, but the best-case scenarios show almost full capacity utilization and a Nash adherence of about 85\%.







\section{Conclusions}\label{sec:conclusions}
This paper presents a multi-agent policy gradient algorithm to explore strategic bidding behavior of shippers and carriers in self-organizing logistics and the Physical Internet. When moving towards automated negotiations and dynamic utilization of transport capacity, bidding strategies have a major impact on the system's stability. In our setting, the agents only observe whether their bids/asks are accepted or rejected, learning and adapting their strategy based on this limited information while their opponent does the same. As strategies may be updated at any point in time, we are dealing with an online non-stationary learning problem.

To model the transport market, we propose a deep reinforcement learning algorithm rooted in policy gradient theory to learn the strategies for bidding and asking respectively. A neutral broker schedules jobs in a way that maximizes the bid-ask spread. This mechanism ensures that the highest bidders and most economical transport services get preference, which would be desirable in real life.

We perform a number of numerical experiments, analyzing the results based on desired properties in bargaining games. Adherence to Nash equilibria, utilization of transport capacity and fairness are defined as key performance metrics. Any solution that divides the market value (difference between maximum willingness to pay and transport costs) between carrier and shipper is a Nash equilibrium, and as such  may be seen as the optimal solution. Approximating the Nash equilibrium is only possible with high utilization. A fair division between of market value between agents is not necessary for an equilibrium, but a sense of fairness is preferable in practical environments. From an individual perspective, the agents attempt solely to maximize their own rewards.

In a deterministic test case, the ideal market situation is approximated quite closely. With both agents adopting comparable learning algorithms, we obtain stable markets that score well on the KPIs, with $\sim$99\% of jobs being shipped and Nash adherence and fairness metrics being well over 90\%. When varying the risk profiles of the agents, we find that being more risk-seeking (in terms of opening bid and ask, penalizing failed negotiations and responsiveness to changes) than the opponent is generally rewarded. This gives agents a strong incentive to strategically place bids and asks. Fairness suffers as a result, although utilization and Nash adherence remain high. 


In a stochastic case, the best attained Nash adherence is $\sim$85\%. As the market is more uncertain, bid-ask spreads increase and the broker takes a larger reward share. We find that cautious strategy updates are necessary to preserve system stability. To improve individual gains, the initialization of the bids and asks is the key determinant. Bold opening bids and asks tend to yield higher rewards, yet if both agents adopt the same strategy results are poor for all included, illustrating the strategic complexities of the environment. 

For both test cases, the ability to improve one's reward share by risk-seeking behavior has a potential drawback from a system perspective. The results show that -- especially if both agents engage in overly risky behavior -- the system performance may strongly decline or even be unstable. Being unable to observe the opponent's strategy, this gives cause for concern when designing fully decentralized transport markets. From a game-theoretical perspective, an argument might be made to avoid risk-seeking behavior to safeguard long-term rewards.

Overall, the results are encouraging as a stepping stone towards decentralized and self-organizing freight transport markets. After some calibration efforts, we consistently obtain solutions that approximate the Nash equilibria describing optimal markets. By appropriately designing their bidding- and asking algorithms, market actors can embed their real-life preferences and risk appetite. The neutral broker appropriately assigns jobs, taking a larger reward share in markets characterized by uncertainty. Generally, we expect to see such properties in a healthy market. To conclude, the evaluated algorithmic design results in a well-functioning and self-organized market without reliance on contracts, regulations or communication protocols.

\bibliographystyle{apa}    
\bibliography{bibliographyphysicalinternet}

\newpage
\section*{Appendix A: Proofs}
\textbf{Lemma 1: Job selection when bid price is lower than ask price}\\
\textit{If $b_{\boldsymbol{j}}<a_{\boldsymbol{j}}$, the broker will always set ${\gamma}_{\boldsymbol{j}}=0$ to maximize its profits.}

\textbf{Proof:}
Take any pair $(b_{\boldsymbol{j}},a_{\boldsymbol{j}}) \in \mathbb{R}^2$ satisfying $b_{\boldsymbol{j}}<a_{\boldsymbol{j}}$. Then ${\gamma}_{\boldsymbol{j}}(b_{\boldsymbol{j}} - a_{\boldsymbol{j}})<0$ if ${\gamma}_{\boldsymbol{j}}=1$ and ${\gamma}_{\boldsymbol{j}}(b_{\boldsymbol{j}} - a_{\boldsymbol{j}})=0$ if ${\gamma}_{\boldsymbol{j}}=0$. As Equation~\eqref{eq:selectionbroker} is a maximization function, it follows that ${\gamma}_{\boldsymbol{j}}=0, \forall b_{\boldsymbol{j}}<a_{\boldsymbol{j}}$. $\blacksquare$

\textbf{Lemma 2: Existence of disagreement point}\\
\textit{For the carrier, there exists a strategy $\pi^C$ that ensures the shipper's payoff $r_{\boldsymbol{j}}^S$ equals at most 0 for any job $\boldsymbol{j}$, regardless of the opponent's strategy $\pi^S$.}

\textbf{Proof:}
Take any job $\boldsymbol{j}$. Let $a_{\boldsymbol{j}} \in \mathbb{R}$ be the ask price. If $c^{C,trn}\leq a_{\boldsymbol{j}}  \leq b_{\boldsymbol{j}}$ is satisfied, the carrier earns $a_{\boldsymbol{j}} -c^{min}\geq 0$ if $\gamma_{\boldsymbol{j}}=1$ and 0 if $\gamma_{\boldsymbol{j}}=0$ (see Equation~\eqref{eq:rewardcarrier}). As both strategies map onto the real domain, there exists pairs of strategies $(\pi^C,\pi^S)$ that ensure nonnegative payoffs for the carrier. The carrier needs not to invoke the disagreement point in this case.

Now suppose $a_{\boldsymbol{j}}  \leq b_{\boldsymbol{j}} < c_{\boldsymbol{j}}^{C,trn}$. In this case, the carrier earns $a_{\boldsymbol{j}} -c_{\boldsymbol{j}}^{C,trn}<0$ if $\gamma_{\boldsymbol{j}}=1$ and can achieve at most 0  (when $\gamma_{\boldsymbol{j}}=0$). It follows that if $b_{\boldsymbol{j}} <c_{\boldsymbol{j}}^{C,trn}$, the carrier is better off asking $a_{\boldsymbol{j}} $ such that $b_{\boldsymbol{j}}<c_{\boldsymbol{j}}^{C,trn}\leq a_{\boldsymbol{j}} $, which by Lemma~1 always guarantees that $\gamma_{\boldsymbol{j}}=0$, yielding a payoff of 0 for both carrier and shipper. To execute the threat, it is not necessary that $b_{\boldsymbol{j}}<c_{\boldsymbol{j}}^{C,trn}\leq a_{\boldsymbol{j}}$ always holds. Following Equation~\eqref{eq:rewardshipper}, if $\pi^C$ yield $a_{\boldsymbol{j}}\geq c_{\boldsymbol{j}}^{S,max}$, the shipper's reward $r_{\boldsymbol{j}}^{S}$ can be at most 0. If $b_{\boldsymbol{j}}\geq a_{\boldsymbol{j}}  \geq c_{\boldsymbol{j}}^{S,max}$, the shipper receives either $c_{\boldsymbol{j}}^{S,max}- b_{\boldsymbol{j}}\leq 0$ or 0. If $b_{\boldsymbol{j}}< a_{\boldsymbol{j}}$, the payoff is by default 0 (Lemma~1). Thus, setting $a_{\boldsymbol{j}} \geq c^{S,max}$ guarantees that the payoff of the shipper is at most 0; a result holding for all bids $b_{\boldsymbol{j}} \in \mathbb{R}$ and thus for all strategies $\pi^S \in \Pi^S$.  $\blacksquare$

\textbf{Lemma 3: Nash equilibrium}
\textit{Any payoff profile satisfying $r_{\boldsymbol{j}}^C+r_{\boldsymbol{j}}^S \equiv c_{\boldsymbol{j}}^{S,max}-c_{\boldsymbol{j}}^{C,trn}$ is a Nash equilibrium. This payoff profile is achieved by any pair of strategies satisfying $c_{\boldsymbol{j}}^{C,trn}\leq a_{\boldsymbol{j}} = b_{\boldsymbol{j}} \leq c_{\boldsymbol{j}}^{S,max}$.}

\textbf{Proof:}
Take any job ${\boldsymbol{j}}$. 
We first show that rational agents will adopt strategies that satisfy $c_{\boldsymbol{j}}^{C,trn} \leq a_{\boldsymbol{j}} \leq b_{\boldsymbol{j}} \leq c_{\boldsymbol{j}}^{S,max}$, being the only profile that may yield positive payoffs. After establishing that this is the feasible set, we continue to prove the Nash equilibrium.

It cannot be guaranteed that $\gamma_{\boldsymbol{j}}=1$ for any bid-ask pair; from Equation~\eqref{eq:selectionbroker} it follows this decision also depends on availability of sufficient capacity. However, $\gamma_{\boldsymbol{j}}=0$ yields a payoff of 0, which is equivalent to the disagreement point. For this reason, we always refer to payoffs $\geq 0$ or $\leq 0$. We distinguish between four sets of payoff profiles.

\textit{\underline{Set 1} (not feasible):} $b_{\boldsymbol{j}}<c_{\boldsymbol{j}}^{C,trn}$\\
The carrier cannot obtain a positive payoff with any $a_{\boldsymbol{j}}$ satisfying $a_{\boldsymbol{j}} \leq b_{\boldsymbol{j}}<c_{\boldsymbol{j}}^{C,trn}$. From Lemma 2, it is therefore beneficial for the carrier to execute the threat, ensuring a payoff of 0 for both agents.

\textit{\underline{Set 2} (not feasible):} $a_{\boldsymbol{j}}>c_{\boldsymbol{j}}^{S,max}$\\
The shipper cannot obtain a positive payoff with any $b_{\boldsymbol{j}}$ satisfying $b_{\boldsymbol{j}} \geq a_{\boldsymbol{j}}>c_{\boldsymbol{j}}^{S,max}$. From Lemma~1, it is beneficial for the shipper to execute the threat, ensuring a payoff of 0 for both agents.

\textit{\underline{Set 3} (not feasible):} $c_{\boldsymbol{j}}^{C,trn} \leq b_{\boldsymbol{j}} < a_{\boldsymbol{j}} \leq c_{\boldsymbol{j}}^{S,max}$\\
From Lemma~1, any bid-ask pair for which $b_{\boldsymbol{j}}<a_{\boldsymbol{j}}$ ensures a payoff of 0 for both agents.


\textit{\underline{Set 4} (feasible):} $c_{\boldsymbol{j}}^{C,trn} \leq a_{\boldsymbol{j}} \leq b_{\boldsymbol{j}} \leq c_{\boldsymbol{j}}^{S,max}$\\
The carrier may earn $a_{\boldsymbol{j}}^\prime - c_{\boldsymbol{j}}^{C,trn}\geq 0$ and the shipper may earn $c_{\boldsymbol{j}}^{S,max}-b_{\boldsymbol{j}}^\prime\geq0$. Both agents have no incentive to invoke the disagreement point as payoffs are greater than or equal to 0.

We established that Set 4 is the only feasible set; positive payoffs can only be attained when adhering to $c_{\boldsymbol{j}}^{trn} \leq a_{\boldsymbol{j}} \leq b_{\boldsymbol{j}} \leq c_{\boldsymbol{j}}^{max}$. We now show that $r_{\boldsymbol{j}}^C+r_{\boldsymbol{j}}^S=c_{\boldsymbol{j}}^{S,max}-c_{\boldsymbol{j}}^{C,trn}$ must be satisfied for a Nash equilibrium. For a proof by contradiction, suppose $c_{\boldsymbol{j}}^{C,trn}\leq a_{\boldsymbol{j}}<b_{\boldsymbol{j}}\leq c_{\boldsymbol{j}}^{S,max}$ holds. The carrier may earn $a_{\boldsymbol{j}} - c_{\boldsymbol{j}}^{C,trn}\geq 0$. However, it could unilaterally improve its payoff by selecting any $a_{\boldsymbol{j}}^\prime$ satisfying $a_{\boldsymbol{j}}<a_{\boldsymbol{j}}^\prime\leq b_{\boldsymbol{j}}$, yielding a potential payoff $a_{\boldsymbol{j}}^\prime - c_{\boldsymbol{j}}^{C,trn}>a_{\boldsymbol{j}} - c_{\boldsymbol{j}}^{C,trn}$ without triggering the disagreement point. Similarly, the shipper may select any $b_{\boldsymbol{j}}^\prime$ satisfying $a_{\boldsymbol{j}} \leq b_{\boldsymbol{j}}^\prime<b_{\boldsymbol{j}}$, potentially yielding $c_{\boldsymbol{j}}^{S,max}-b_{\boldsymbol{j}}^\prime>c^{S,max}-b_{\boldsymbol{j}}$ without triggering the disagreement point. It follows that $c_{\boldsymbol{j}}^{C,trn}\leq a_{\boldsymbol{j}} = b_{\boldsymbol{j}} \leq c_{\boldsymbol{j}}^{S,max}$ is the equilibrium point. Any unilateral deviation would either reduce the agent's own payoff or invoke the disagreement point. Under the condition that $a_{\boldsymbol{j}}=b_{\boldsymbol{j}}$, subject to $c_{\boldsymbol{j}}^{C,trn}\leq a_{\boldsymbol{j}}$ and $b_{\boldsymbol{j}} \leq c_{\boldsymbol{j}}^{S,max}$, the payoff is  $(a_{\boldsymbol{j}} - c_{\boldsymbol{j}}^{C,trn})+(c_{\boldsymbol{j}}^{S,max}-b_{\boldsymbol{j}})=(a_{\boldsymbol{j}} - c_{\boldsymbol{j}}^{C,trn})+(c_{\boldsymbol{j}}^{S,max}-a_{\boldsymbol{j}})=c_{\boldsymbol{j}}^{S,max}- c_{\boldsymbol{j}}^{C,trn}$; exactly the full system gain is divided between the agents. This result is equivalent to the equilibrium for the bargaining game as defined by Nash \cite{nash1953}.  $\blacksquare$

\newpage

\section*{Appendix B: Experimental results for actor network architectures}
This appendix shows the results of various actor network architectures, corresponding to Section~\ref{sssec:actor_networks_architectures}. See Table~\ref{table:network_architectures_appendix} for the results per network architecture.
	
	\begin{table}[h!]
		\scriptsize	
	\centering
	\caption{Comparison of various actor network architectures. In general, we observe worse average results for deep networks and worse average results for large numbers of nodes. End-of-horizon values are generally better than for smaller networks.}
	\label{table:network_architectures_appendix}
	\begin{tabular}{ l l l l  }
		\toprule
		Architecture & Utilization &  Nash adherence & Fairness \\
		\midrule	
		Linear & 0.99 (1.00) &0.96 (0.97)&0.86 (0.97)  \\
		
		1L5N & 0.99 (1.00) &0.93 (0.95)&0.64 (0.85)  \\
		1L10N & 0.99 (0.99)&0.92 (0.92)& 0.93 (0.93) \\
		1L15N & 0.99 (0.99)&0.91 (0.92)& 0.85 (0.94) \\
		1L20N & 0.99 (1.00) &0.91 (0.91)& 0.95 (0.95) \\
		1L25N & 0.99 (0.99) &0.91 (0.89)& 0.92 (0.96) \\
		1L30N & 0.99 (0.99) &0.90 (0.92)& 0.94 (0.96)\\
		
		2L5N & 0.99 (0.99) &0.90 (0.93)&0.60 (0.93)  \\
		2L10N & 0.99 (0.99)&0.88 (0.91)& 0.91 (0.94) \\
		2L15N & 0.99 (0.99)&0.89 (0.90)& 0.93 (0.94) \\
		2L20N & 0.99 (0.99) &0.89 (0.91)& 0.92 (0.94) \\
		2L25N & 0.99 (0.99) &0.89 (0.93)& 0.91 (0.90) \\
		2L30N & 0.87 (0.99) &0.76 (0.91)& 0.60 (0.89)\\
		
		3L5N & 0.99 (0.99)  &0.91 (0.92)&0.88 (0.86)  \\
		3L10N & 0.99 (0.99) &0.89 (0.94)& 0.94 (0.94) \\
		3L15N & 0.99 (0.99) &0.89 (0.94)& 0.94 (0.94) \\
		3L20N & 0.99 (0.99) &0.89 (0.93)& 0.92 (0.96) \\
		3L25N & 0.74 (1.00) &0.61 (0.92)& 0.22 (0.35) \\
		3L30N & 0.94 (0.99) &0.85 (0.94)& 0.82 (0.88)\\
		\bottomrule
	\end{tabular}
\end{table}

\newpage
\section*{Appendix C: Bias initialization}
At the start of each experiment, we initialize the bias weight, with the bias node itself being a node with value 1. With all other weights at the output layer being initialized at 0, this means that the initial bids and asks are equal to the bias weights provided. Informally the bias initialization can be seen as the opening bid or ask, greatly influencing the eventual convergence.

The costs and willingness to pay for each job depend on two job characteristics: distance and volume. Let
$\mathbf{d}=[1,\ldots,d^{max}]$ be the vector of feasible distances and $\mathbf{v}=[1,\ldots,\zeta^{max}]$ be the vector of feasible volumes. By taking the outer product of the vectors, we obtain a matrix containing the marginal weight of each job type:
 
\begin{align}
{\displaystyle \mathbf{d} \otimes \mathbf {v} ={\begin{bmatrix}d_{1}v_{1}&d_{1}v_{2}&\dots &d_{1}v_{\zeta^{max}}\\d_{2}v_{1}&d_{2}v_{2}&\dots &d_{2}v_{\zeta^{max}}\\\vdots &\vdots &\ddots &\vdots \\d_{d^{max}}v_{1}&d_{d^{max}}v_{2}&\dots &d_{d^{max}}v_{\zeta^{max}}\end{bmatrix}}}\enspace.\notag
\end{align}
 
Recall that ${c}^{C,trn}$ describes the transport cost per volume unit per distance unit and ${c}^{S,max}$ the willingness to pay. The transport cost matrix -- containing transport costs for each job type -- is then given by $\left(\mathbf{d} \otimes \mathbf {v}\right) \cdot c^C$. Similarly, we may retrieve the willingness to pay for each job type by computing  $\left(\mathbf{d} \otimes \mathbf {v}\right) \cdot {c}^{S,max}$. As we assume independent uniform probability distributions for every job characteristic, each matrix element has an equal probability. Thus, the average transport costs and average willingness to pay are simple the averages of all elements in the respective matrices.

We use the average transport cost $\bar{c}^{C,trn}$ and average willingness to pay $\bar{c}^{S,max}$ to initialize the bias weights. To provide some intuition: if the shipper would set $\bar{c}^{C,trn}$ as the initial bias weight, it would on average pose bids equal to the average transport costs for all jobs. As we use a single actor network for all jobs, the bias is a constant that holds for all job types. At the risk of being superfluous, we stress that the network must still learn the values corresponding to the various job types.

The initial bias weights are formalized in Table~\ref{table:bias initializations}.

\begin{table}[h!]
	\scriptsize
		\centering
	\caption{Bias initializations per risk profile.}
	\label{table:bias initializations}      
	\begin{tabular}{ l l l}
		\toprule
		Risk profile & Carrier &  Shipper \\
		\midrule	
		Risk-seeking & $\bar{c}^{S,max}$ & $\bar{c}^{C,trn}$\\
		Risk-neutral &  $(\bar{c}^{C,trn}+\bar{c}^{S,max})/2$&  $(\bar{c}^{C,trn}+\bar{c}^{S,max})/2$ \\
		Risk-averse & $\bar{c}^{C,trn}$ & $\bar{c}^{S,max}$ \\
		\bottomrule  
	\end{tabular}
\end{table}

\newpage
\section*{Appendix D: Nash equilibria}
This appendix investigates the existence of pure Nash equilibria when agents adopt asymmetric strategies. Assuming the opponent's strategy is fixed, an agent can determine its best response to that strategy. When both agents are cannot improve rewards by unilaterally changing their strategy, an equilibrium is reached. There may be multiple equilibria. The results -- for Case I, Case II ($\zeta=40$) and Case II ($\zeta=300$) respectively -- are found in Table~\ref{table:nash equilibria case I}, Table~\ref{table:nash equilibria case II (40cap)} and Table~\ref{table:nash equilibria case II (300cap)}. The tables contain the normalized net rewards per agent (which depend on both the reward shares and the number of jobs shipped) and the best responses per agent.

	\begin{table}[h!]
		\scriptsize
		\centering
		\caption{Normalized net rewards per agent. Best individual responses (Case I) to an opponent's strategy are \underline{underlined}, Nash equilibria are in \textbf{bold}.}
		\label{table:nash equilibria case I}
		\begin{tabular}{ l| l l l  }
			\hline
			\backslashbox{Carrier}{Shipper} & RS &  RN & RA \\
			\hline		
			RS &	0.00/0.00  &	\underline{\textbf{0.69}}/\underline{\textbf{0.15}} &	\underline{0.78}/0.13\\
			RN &	0.24/\underline{0.73} &	0.46/0.49 &	0.61/0.28\\
			RA &	\underline{\textbf{0.80}}/\underline{\textbf{0.83}}  &	0.26/0.62 &	0.41/0.40\\
			\hline
		\end{tabular}
	\end{table}

\begin{table}[h!]
	\scriptsize
	\centering
	\caption{Normalized net rewards per agent. Best individual responses (Case II, $\zeta=40$) to an opponent's strategy are \underline{underlined}, Nash equilibria are in \textbf{bold}.}
	\label{table:nash equilibria case II (40cap)}
	\begin{tabular}{ l| l l l  }
		\hline
		\backslashbox{Carrier}{Shipper} & RS Bias &  RN Bias & RA Bias \\
		\midrule		
		RS Bias&	0.00/0.00  &	0.15/0.57 &	0.08/\underline{0.67}\\
		RN Bias&	\underline{\textbf{0.01}}/\underline{\textbf{0.57}} &	\underline{0.39}/0.53 &	\underline{0.87}/-0.01\\
		RA Bias&	0.00/\underline{0.90}  &	0.32/0.47 &	0.66/0.05\\
		\hline
	\end{tabular}
\end{table}	

\begin{table}[h!]
	\scriptsize
	\centering
	\caption{Normalized net rewards per agent. Best individual responses  (Case II, $\zeta=300$)  to an opponent's strategy are \underline{underlined}, Nash equilibria are in \textbf{bold}.}
	\label{table:nash equilibria case II (300cap)}
	\begin{tabular}{ l| l l l  }
		\hline
		\backslashbox{Carrier}{Shipper}  & RS Bias&  RN Bias & RA Bias\\
		\hline		
		RS Bias&	0.00/0.11  &	0.08/\underline{0.67} &	0.68/0.21\\
		RN Bias&	-0.07/\underline{0.76} &	\underline{0.31}/0.59 &	\underline{0.94}/-0.08\\
		RA Bias&	\underline{\textbf{0.01}}/\underline{\textbf{0.87}}  &	0.29/0.49 &	0.49/0.04\\
		\hline
	\end{tabular}
\end{table}

\end{document}